\documentclass[conference]{IEEEtran}
\IEEEoverridecommandlockouts
\usepackage{cite}
\usepackage{amsmath,amssymb,amsfonts}
\usepackage{algorithmic}
\usepackage{graphicx}
\usepackage{textcomp}
\usepackage[dvipsnames,svgnames,x11names]{xcolor}
\def\BibTeX{{\rm B\kern-.05em{\sc i\kern-.025em b}\kern-.08em
    T\kern-.1667em\lower.7ex\hbox{E}\kern-.125emX}}

\usepackage{bm} 
\usepackage{booktabs} 
\usepackage{subfigure} 
\usepackage{amsmath} 
\usepackage{multirow} 
\usepackage{makecell} 
\usepackage{url} 
\usepackage{enumitem} 
\usepackage[ruled,linesnumbered]{algorithm2e} 
\usepackage{xspace} 

\usepackage{todonotes} 
\newcommand{\hide}[1]{} 

\newcommand{\xiao}[1]{#1}
\newcommand{\xcx}[1]{{\color{black}#1}}
\newcommand{\zhou}[1]{{\color{black}#1}}
\newcommand{\zhoure}[1]{{\color{black}#1}}

\newcommand{\zjb}[1]{{\color{black}#1}}
\newcommand{\xcxre}[1]{{\color{black}#1}}
\newcommand{\old}[1]{}

\newcommand{\mymodel}{{CMSF}\xspace}

\newcommand{\compb}{{MAGA}\xspace}
\newcommand{\compc}{{GSCM}\xspace}
\newcommand{\compd}{{MS-Gate}\xspace}

\begin{document}

\title{A Contextual Master-Slave Framework on \\ Urban Region Graph 
for Urban Village Detection}

\author{\IEEEauthorblockN{Congxi Xiao$^{1,2,\dagger}$, Jingbo Zhou$^{2*}$, Jizhou Huang$^{3}$, Hengshu Zhu$^{3}$, Tong Xu$^{1}$, Dejing Dou$^{2*}$, Hui Xiong$^{4,5*}$
\thanks{$^\dagger$This work was done when the first author was an intern in Baidu Research under the supervision of Jingbo Zhou.}
\thanks{$*$Corresponding authors.}}
\IEEEauthorblockA{
\textit{$^{1}$University of Science and Technology of China},
\textit{$^{2}$Business Intelligence Lab, Baidu Research}, \\
\textit{$^{3}$Baidu Inc.},
\textit{$^{4}$The Hong Kong University of Science and Technology (Guangzhou)},\\
\textit{$^{5}$Guangzhou HKUST Fok Ying Tung Research Institute}.\\
\{xiaocongxi, zhoujingbo, huangjizhou01, zhuhengshu\}@baidu.com, \\tongxu@ustc.edu.cn, dejingdou@gmail.com, xionghui@ust.hk}}

\maketitle
\begin{abstract}
Urban villages (UVs) refer to the underdeveloped informal settlement falling behind the rapid urbanization in a city. Since there are high levels of social inequality and social risks in these UVs, it is critical for city managers to discover all UVs for making appropriate renovation policies. Existing approaches to detecting UVs are labor-intensive or have not fully addressed the unique challenges in UV detection such as the scarcity of labeled UVs and the diverse urban patterns in different regions. 
To this end, we first build an urban region graph (URG) to model the urban area in a hierarchically structured way. Then, we design a novel contextual master-slave framework to effectively detect the urban village from the URG. The core idea of such a framework is to firstly pre-train a basis (or master) model over the URG, and then to adaptively derive specific (or slave) models from the basis model for different regions. The proposed framework can learn to balance the generality and specificity for UV detection in an urban area. Finally, we conduct extensive experiments in three cities to demonstrate the effectiveness of our approach. 
\end{abstract}

\begin{IEEEkeywords}
Urban Villages, Urban Region Graph, Graph Neural Networks, Master-Slave Framework.
\end{IEEEkeywords}

\section{Introduction}
The imbalanced development of urban cities has led to the formation of urban villages (UVs) due to the absence of planning and management. 
Because of the low house rent, UVs have gradually become settlements of migrants and low-income groups who also have significant socioeconomic contributions to the city. However, UVs face serious social inequality and social risks, such as high epidemic infection risks \hide{\cite{xiao2021c, ren2019urban}}\cite{ren2019urban}, harmful environmental pollution \cite{huang2015spatiotemporal,liu2017use}, and 
poor public order \cite{brindley2003social}.
According to the Sustainable Development Goals  Report 2018 \cite{sdg2018sustainable},  more than 1 billion people live in such UV-like informal settlements worldwide. Good knowledge of UVs will help city planners to make appropriate renovation policies aiming to build sustainable cities and communities. 

As the first step towards solving the UV problem, how to detect UVs in a whole city is recognized as an indispensable but challenging task. The city planner usually fails to possess the  panorama of UV distribution, and even local authorities of various levels only know a fairly limited part of them (e.g. some well-known communities).
The traditional detection methods mainly depend on fieldwork and social investigation by the government \cite{pan2020deep,huang2015spatiotemporal}, which are impracticably time-consuming and labor-intensive. 
Therefore, a lot of research attention \cite{huang2015spatiotemporal,shi2019domain,mast2020mapping,zhao2020partition,pan2020deep,chen2021uvlens,chen2022hierarchical} has lain in detecting UVs in the city based on data-driven learning with some expert annotations.
The early index-based approaches \cite{huang2015spatiotemporal, liu2017use} use classic machine learning models with a series of hand-crafted metrics from satellite images for UV detection. Some studies also attempt to incorporate multi-modal data \cite{zhao2020partition,chen2021uvlens,chen2022hierarchical} for UV detection.

There are a few recent studies \cite{mast2020mapping,pan2020deep,chen2019identifying} to handle UV detection by directly adopting advanced deep learning (DL) models, such as fully convolutional neural networks \cite{mast2020mapping}, U-Net \cite{pan2020deep} and Mask-RCNN \cite{chen2019identifying}, but without considering the special challenging points of this problem.
Nevertheless, we observe that there are two unique issues for UV detection which have not been seriously investigated by previous studies. First of all, the scarcity of labeled UVs can severely undermine the recognition capacity of DL models. UVs take only a very minor part of the urban regions, and the number of labeled UVs in a city is even much smaller. Whereas, existing DL models usually require sufficient labeled data to obtain satisfactory generalization abilities \cite{wan2021contrastive}.
Second, there is diversity in the urban area with different characteristics and data distribution (e.g. downtown vs suburb). It is hard to train a single model for UV detection that works well across all urban areas with diverse patterns. This problem becomes even more serious considering the scarcity of labeled UVs.  

To this end, we propose a Contextual Master-Slave Framework, named as \textbf{\mymodel}, tailored for the UV detection problem. \mymodel is running on an Urban Region Graph (URG) carefully built for modeling the urban area.  The general idea of the  \mymodel is to pre-train a basis (or master) model first on the URG, and then derive a specific (i.e. slave) model for each region given the contextual information learned from the URG.

The URG takes fine-grained regions as nodes, and builds region relations in the urban area as edges according to regions' spatial context and road network connectivity, which uncover the geographical and functional correlations among regions. Moreover, a set of region features are extracted from Point of Interest (POI) data and satellite image data to reflect the infrastructure distribution and visual region appearance, respectively. Modeling such socioeconomic conditions of regions plays a critical role in urban village detection.

To be specific, based on the URG, \mymodel is trained in two stages: 1) taking full knowledge of limited labeled UVs to pre-train a basis (or master) model over the URG, and also to learn region representation; and 2) learning to derive specific (or slave) models for different regions by leveraging the context information to moderate the pre-trained master model.

In the first stage, we pre-train a specially designed hierarchical graph neural network as the master model and learn the region representation during the training process. At first, in view of the complementarity between different modal data (POIs and images) of the URG, we design a Mutual-Attentive Graph Aggregation layer (\compb) to leverage the inter-modal context for region representation enhancement. Second, we design a global semantic clustering method (\compc) to cluster semantically similar regions together based on their representation learned from region features and organize the urban area as a hierarchical structure.
This structure enables clusters to capture the global semantic context from distant but similar regions inside, through performing message collection from similar regions in $regions \rightarrow clusters$ direction, and then propagate this shared knowledge back in $clusters \rightarrow regions$ direction. In this way, the region representation can also extract the region's contextual information from the URG. 

In the second stage, we devise a novel contextual master-slave gating mechanism (\compd) to learn a gate function that can help to adaptively derive a slave model given each region's contextual information.
Considering the diverse patterns of UVs in different urban areas (for example, the UV in downtown might be different from the one in suburb), it is better to train a specific model for a certain cluster of regions.
However, due to the scarcity of known UVs in a city, many clusters will have too few UVs to effectively train their individual models.
To balance the generality and specificity, during the second training stage, we optimize a gate function which can encode contextual information of each region and moderate the master model to adaptively derive a region-wise slave model. The master model is also jointly fine-tuned in this stage. After this two-stage training, we can derive a specific predictor for each region to achieve more accurate UV detection.

We conducted extensive evaluations for UV detection in three cities in China. Experimental results show that our framework can achieve significant improvement over other state-of-the-art methods on several metrics, including Area Under Curve (AUC), Precision, Recall, and F1-score.
Our contributions can be summarized as follows:
\begin{itemize}[leftmargin=*]
    \setlength{\itemsep}{3pt}
    \item We propose \mymodel, a novel contextual master-slave framework to handle the unique challenges in the UV detection problem. Instead of training a single model over the city, \mymodel aims to utilize the region's contextual information to adaptively derive a specific predictor for each region.
    \item \xcxre{To the best of our knowledge, we are the first to study the UV detection problem from a graph perspective. We construct an URG based on multiple sources of urban data, to model the dependencies among regions. Upon the URG, we design a hierarchical graph neural network which can make full use of limited labeled UVs through global semantic clustering, to effectively learn the region representation and extract rich context information for deriving slave models.}
    \item We conducted extensive experiments in three cities in China to demonstrate the superior detecting ability of \mymodel.
\end{itemize}

\section{RELATED WORK}

Urban village detection has attracted a lot of research attention from the \zjb{data mining and} geoscience community. 
With the increasing availability of high-resolution satellite images, some index-based approaches  \cite{huang2015spatiotemporal, liu2017use} are devised to use classic machine learning models for classification upon hand-crafted metrics from images, such as the mean of RGB and MBI index \cite{huang2011multidirectional}. In recent years, several studies \zjb{try to} handle the UV detection problem by 
building different deep learning models 
over satellite images \cite{li2017unsupervised, pan2020deep,mast2020mapping,shi2019domain}.  Considering the limitation of the single image perspective, \zjb{there are also a few recent studies to integrate }additional data (like taxi trajectories and POIs) with satellite image data to benefit UV detection \cite{chen2019identifying,chen2021uvlens}. 
However, existing studies ignore two important challenges for UV detection: 1) the limited 
number of labeled UVs; and 2) the diverse urban pattern in a city. Our framework \mymodel is specially designed to tackle the above challenges, leading to much better performance than existing solutions.

Note that a similar concept named master-slave regularized model is investigated by \cite{xu2020adaptive}. But their objective is to use a master model to directly predict model parameters of logistic regression models for company revenue prediction, whose methodology and application domain are  different from ours.
\zhou{Another close concept is the semi-lazy learning method \cite{zhou2015smiler,zhou2013semi,zhou2013r2} which tries to build an individual model for each instance upon nearest (or similar) neighbors. This method usually has a high cost to build the model online after retrieving data. Therefore, usually it is not suitable to adopt this methodology with the deep learning method.
}
\xcxre{Other works like \cite{yao2019learning} adopting meta-optimization for cross-city urban applications seems similar with the idea of master-slave model. An important difference is that such a meta-optimization method first fine-tunes a pre-trained initial model to different datasets (from different cities), then it is fixed for all input instances in one dataset. However, CMSF is designed to be capable of deriving the slave model for each prediction instance given a region-specific context, making our model different from it.}

Our study is also related to the GNN for urban applications. Much attention has been devoted to applying GNNs (e.g. GCN \cite{kipf2016semi} and GAT \cite{velivckovic2018graph}) for many urban applications, such as region embedding \cite{jenkins2019unsupervised,fu2019efficient}, regional economy prediction \cite{xu2020attentional}, crowd flow forecasting \cite{xia20213dgcn}, traffic demand forecasting \cite{geng2019spatiotemporal}, and real estate appraisal \cite{zhang2021mugrep}.
There are also a few studies to model the city in a hierarchical graph structure for transport-related applications, such as HRNR \cite{wu2020learning} which models the hierarchical road networks for road segment classification and route planning, and STRN \cite{liang2021fine} which partitions the fine-grained urban grid map into a coarse-grained level for urban flow prediction. However, these methods cannot be directly adopted to model the region dependency for UV detection. 

\section{Preliminaries}
In this section, we first introduce the basic concepts and notations used throughout this paper, and then we formally formulate the problem of urban village detection.

Given an urban area of interest (typically it can be the main urban area of a city), we can divide it into $N = H \times W$ non-overlapping \emph{region grids} with a fixed size. 
Hereafter, if without specification, we use \textit{region} to refer to the region grid for convenience.
We use $\mathcal{V}=\{v_1, ..., v_N\}$ to denote a set of regions. 
A Point of Interest (POI) is a specific point location on the map that can provide some useful services. 
Each region $v_i \in \mathcal{V}$ in the urban area usually contains a set of various POIs and is covered by a satellite image showing the appearance of this region.
Upon the POI and satellite image data, we can extract discriminative features $\bm{x}_i \in \mathbb{R}^{d}$ for each region, which are useful for urban village detection. 
We use $\bm{x}^P_i$ and $\bm{x}^I_i$ to denote the constructed POI and image features of region $v_i$ respectively, i.e. $\bm{x}_i=\bm{x}^P_i\cup \bm{x}^I_i$.
How to extract such features will be introduced in Section \ref{features}.

\newtheorem{definition}{Definition}
\begin{definition}
{\bfseries Urban Village Detection.}
Given the partitioned region grids, the urban village detection problem can be defined as a region-wise binary classification task based on the region features: $f(\bm{x}_i) \rightarrow y_i$, where $y_i$ is the binary label indicating that region $v_i$ is contained by or overlapped with an urban village ($y_i=1$) or not ($y_i=0$). 

In our experimental evaluation, the significant overlap is defined as the region and the urban village having an overlap larger than 20\% of the region's area. Note that only a few regions in the city are known to be urban villages. The challenge of this problem is how to associate each unlabeled region with a binary label upon the limited labeled data.
Formally, the region set $\mathcal{V}$ of size $N$ consists of two parts: the labeled region set $\mathcal{V}^L = \{v_1, ...,v_l \}$ with feature matrix $\bm{X}^L \in \mathbb{R}^{l \times d}$ and label matrix $\bm{Y}^L \in \mathbb{R}^l$, and the unlabeled region set $\mathcal{V}^U = \{v_{l+1}, ...,v_N \}$ with feature matrix $\bm{X}^U \in \mathbb{R}^{(N-l) \times d}$. Our goal is to learn a predictive function $f:(\bm{X}^U |\,\bm{X}^L, \bm{Y}^L) \rightarrow \bm{Y}^U$.
\end{definition}

\section{Urban Region Graph}
The URG is defined as $\mathcal{G}(\mathcal{V},\mathcal{E},\bm{A},\bm{X})$, where $\mathcal{V}=\mathcal{V}^L \cup \mathcal{V}^U$ denotes the node set containing all the regions. $\bm{X} \in \mathbb{R}^{N\times d}$ is the corresponding region feature matrix obtained from POI and image data, where $N$ is the number of regions and $d$ is the feature dimension.
$\mathcal{E}$ is the edge set modeling the relation among different regions built from regions' spatial context and the road network of the city.
$\bm{A}$ denotes the adjacency matrix of URG depending on $\mathcal{E}$, where $\bm{A}_{ij} = 1$ if there exists an edge between $v_i$ and $v_j$, otherwise $\bm{A}_{ij} = 0$. 

\begin{figure}[t]
\vspace{-2mm}
  \centering
  \includegraphics[width=0.75\columnwidth]{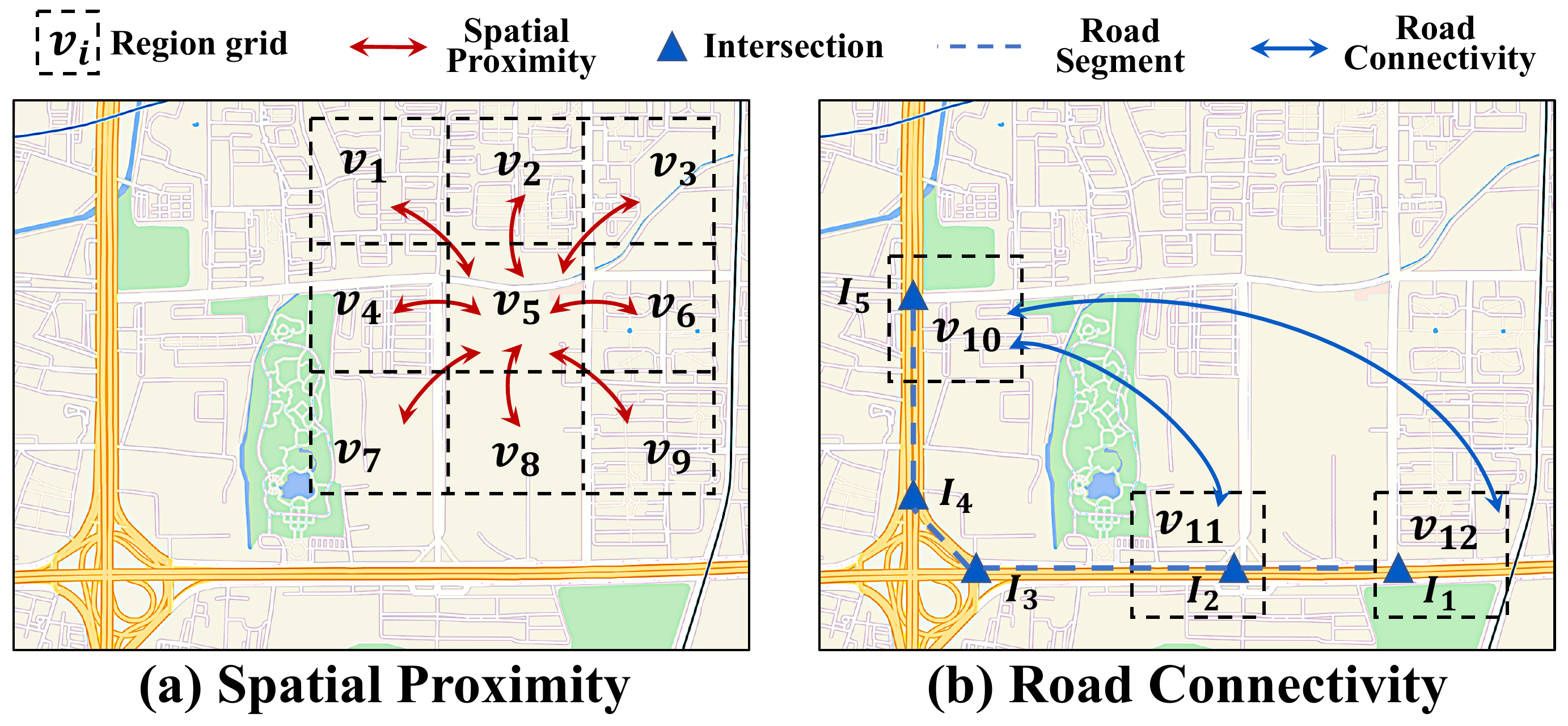}
  \vspace{-2mm}
  \caption{Illustration of Region Relation Construction}
  \label{fig-edge-construct}
  \vspace{-3.5mm}
\end{figure}

\begin{figure}[t]
  \centering
  \includegraphics[width=0.75\columnwidth]{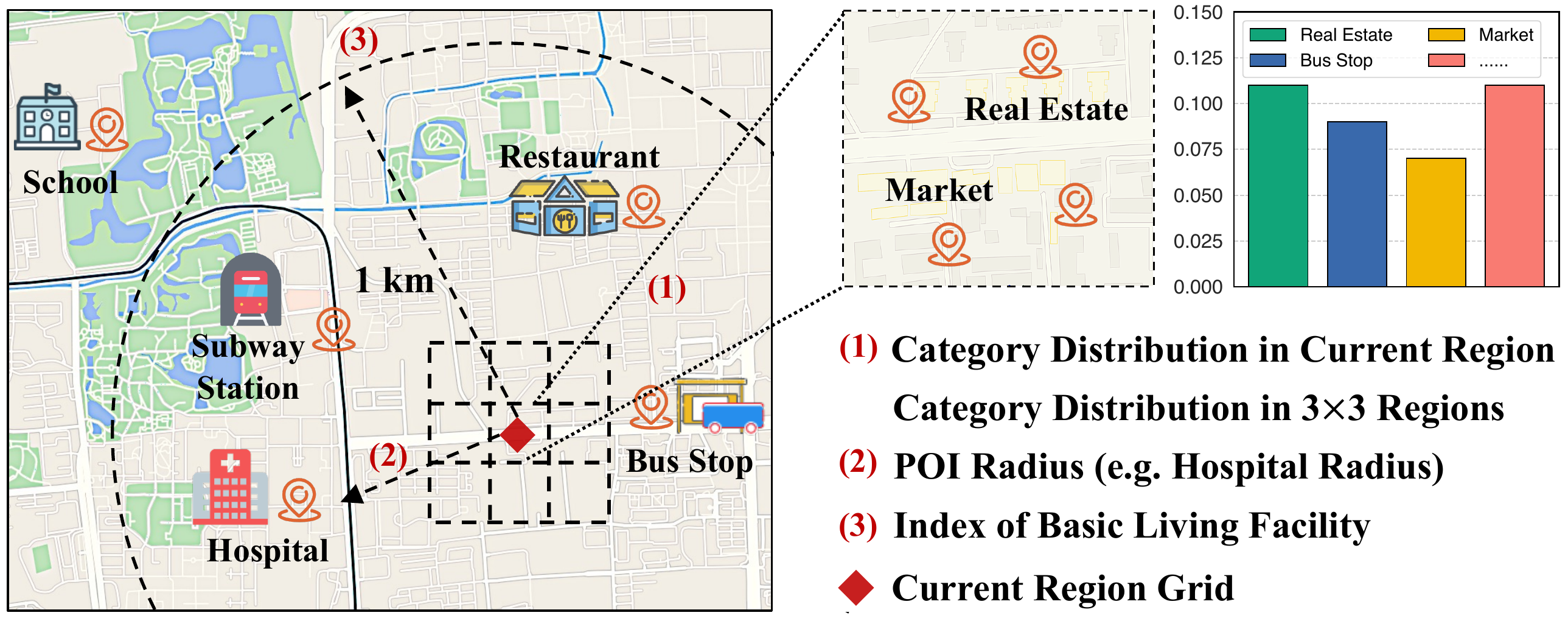}
  \vspace{-1mm}
  \caption{Illustration of Feature Construction}
  \label{fig-features-construct}
  \vspace{-4mm}
\end{figure}

\subsection{Region Relation Construction}
\label{edge-construction}
To model the relation among regions, we collectively build the edge set $\mathcal{E}$ and adjacency matrix $\bm{A}$ of URG from two perspectives, which are \textit{spatial proximity} and \textit{road connectivity}.

\textbf{Spatial Proximity.}
Following the principle of spatial dependence that ``everything is related to everything else, but near things are more related than distant things'' \cite{tobler1970computer}, we consider each region spatially correlated with its neighbors in the grid map, and assume they should have more similar semantics (and thus similar representations in latent space). This assumption is also supported by recent studies modeling the urban area as a grid map \cite{yao2018deep,geng2019spatiotemporal,jean2019tile2vec}.
Therefore, we connect region $v_i$ and $v_j$ if they are mutually one of the eight neighbors to each other in the $3 \times 3$ region grids, and set the corresponding $\bm{A}_{ij} = 1$\hide{ if $|h_i-h_j|+|w_i-w_j| = 1$}. As illustrated in Figure \ref{fig-edge-construct}(a), region $v_5$ will be connected with surrounding regions $v_{1-4}$ and $v_{6-9}$.

\textbf{Road Connectivity.}
\label{road}
In addition to the plain dependency reflected by the adjacent location, road networks can reveal more complex correlations among regions in an urban area.
As the core component of the urban transportation system, existing studies have demonstrated that road networks take an indispensable role in discovering the functionality across regions \cite{wu2020learning}.
Intuitively, a function area can be formulated across regions connected by roads although these regions are not geographically close.
Additionally capturing this function-aware correlation is conducive to complex urban structure modeling. 
Therefore, we incorporate the road network data provided by \cite{karduni2016protocol} to set $\bm{A}_{ij} = 1$ if regions $v_i$ and $v_j$ are mutually connected by roads. 
The road connectivity between two regions is determined by whether they can be reached within a limited number of hops in the road networks (in our experiment, we set regions to be connected within 5 hops). 

Figure \ref{fig-edge-construct}(b) provides an explanatory example of road connectivity between regions to better clarify the detailed rule of building such a relation. The road network data can be treated as a graph, where the nodes represent intersections on the road networks and the edges are road segments that connect the nodes. As shown in Figure \ref{fig-edge-construct}(b), we use triangle and dash line in blue to denote the nodes ($I_1 \sim I_5$) and edges on road networks, respectively. Each node has a unique geographic coordinate (i.e. longitude and latitude), based on which we associate the node to the region grid that it locates in. For example, region $v_{10}$ and $v_{11}$ have node $I_5$ and $I_2$ inside, respectively. Then, we define that region $v_i$ and region $v_j$ are mutually connected by roads if there exists a path containing no more than 5 edges (road segments) between any node in $v_i$ and the one in $v_j$. For example, the region $v_{10}$ and $v_{11}$ are connected by road since they can reach each other with 3 road segments ($I_2 - I_3 - I_4 - I_5$). Based on this rule, we link $v_{10}$ and $v_{11}$ on URG and set $A_{10,11}=A_{11,10}=1$.

\subsection{Feature Construction}
\label{features}
In our method, we mainly use two groups of features to characterize a region, which are POI features and image features. We first briefly introduce the data source and then explain how to extract these features.

The region features are constructed based on the POI basic property data and the satellite image data from Baidu Maps.
POI basic property data provide the name, location, multi-level categories (e.g. \xiao{transportation facility and bus stop}), and other basic information of a POI \cite{xiao2021c, yuan2020spatio}. 
With this POI data, we can describe the human activities and distribution of functional facilities in an urban region \cite{xu2020new}.
The satellite image data used in our paper are 3-channel RGB images with $256 \times 256$ pixels, depicting the appearance of each $128m \times 128m$ region grid in the top view. Their spatial resolution is 0.5 meters. 
We introduce how to extract POI and image features based on these data as follows.

\textbf{POI Features.}
\label{poi-features}
The motivation of extracting features from POI data is that UVs are usually residential areas with substandard living conditions and insufficient basic facilities (e.g. cultural, sports and leisure facilities), which can be justly reflected by POIs in these regions.
Hence, we design the following three types of POI features which are:
\begin{itemize}[leftmargin=*]
    \item \textit{Category Distribution.} 
    We make statistics of POIs belonging to different categories (e.g. catering and life service) in a given region. Then, a distribution histogram vector can be calculated in which the value of each element equals the ratio of the corresponding category. 
    Besides, the total number of POIs in the region is also directly recorded in the feature vector. Note that we additionally calculate the category distribution in the $3 \times 3$ grids centered by the given region to include more surrounding information. 
    \item \textit{POI Radius.}
    For roughly measuring how convenient to access various basic living facilities from a region, we compute a number of different radius features, each of which is defined as the shortest distance between the current region and one type of POIs (e.g. radius to hospital). 
    Note that we categorize the distance into different buckets ($<0.5km$, $0.5\sim1.5km$, $1.5\sim3km$ and $>3km$) for discretization.
    \item \textit{Index of Basic Living Facility.}
    To measure the perfect degree of basic living facilities (e.g. bus stop, hospital, and restaurant), we further define a binary index which is assigned one if a set of living facilities are all within $1km$ of the region, otherwise it is assigned zero. 
\end{itemize}
Finally, the comprehensive POI features are obtained through the concatenation of the three types of features. An illustration of POI features construction for better understanding is shown in Figure \ref{fig-features-construct}, \xcxre{and all the specific POI types considered in our work are listed in Appendix I-B.}

\textbf{Image Features.}
\label{image-features}
We incorporate the satellite imagery information to represent the appearance characteristics of regions. There are also a few studies \cite{chen2021uvlens, shi2019domain} capturing visual features from satellite images to locate UVs, since UVs are usually presented with overcrowded and irregularly arranged buildings as well as narrow alleys in appearance. 

Considering that directly inputting high dimensional pixel-level image data to train the detection model with limited labels will lead to overfitting, we use the VGG16 \cite{simonyan2014very} model pre-trained on ImageNet as a feature extractor to obtain the semantic representation of the satellite image.
Specifically, the raw image data of each region is fed into the pre-trained VGG16 model with the top two fully connected layers removed, then the model outputs the 4096-dimensional vector as image features of each region.

\section{MODEL FRAMEWORK}
\label{method}
\begin{figure*}[t]
\centering
\includegraphics[width=0.9\textwidth]{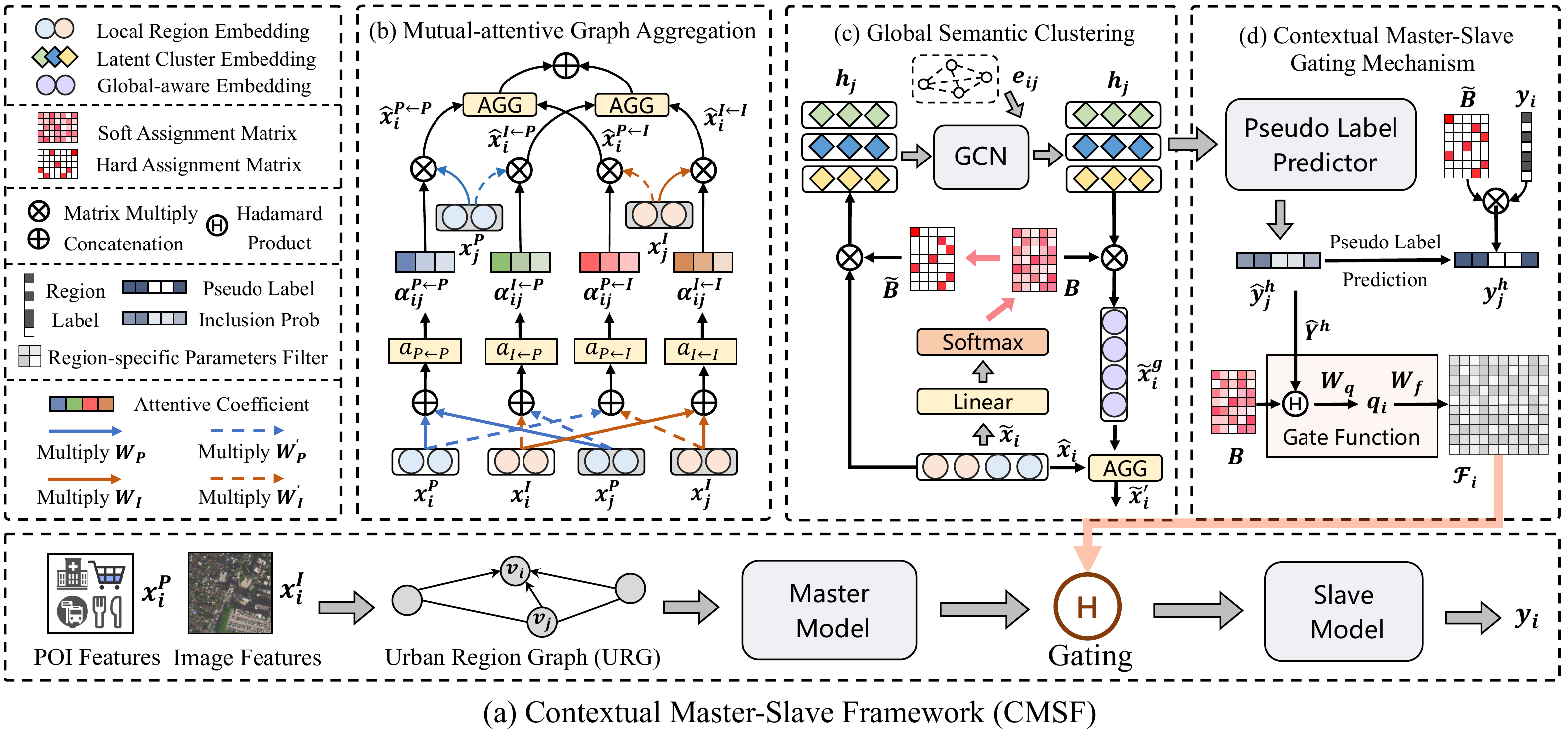}
\vspace{-3mm}
\caption{Illustration of the proposed contextual master-slave framework (\mymodel).}
\vspace{-3mm}
\label{fig_model_framework}
\end{figure*}

\mymodel has two training stages which are 1) the master training stage; and 2) the slave adaptive training stage. An overview of \mymodel is shown in Figure \ref{fig_model_framework}(a). 
During the master training stage, we build a hierarchical graph neural network to learn the region representation. To be specific, we design a mutual-attentive graph aggregation layer (\textit{\compb}, see Figure \ref{fig_model_framework}(b)), and a global semantic clustering module (\textit{\compc}, see Figure \ref{fig_model_framework}(c)) to learn the region representation jointly with local structure context and global semantic context in the urban area.
First, \compb learns enhanced multi-modal region representation from POI and image features upon URG.
Then, \compc is built to globally cluster the semantically similar regions based on the aggregated features from \compb and form a hierarchical structure of the urban area. Through this structure, clusters collect the representation of regions inside as the global semantic context and propagate it back to update the region representation. In this stage, we treat the hierarchical GNN as the master model and pre-train it on the whole city to recognize UVs and learn the  region representation.

In the slave adaptive training stage, we propose a contextual master-slave gating mechanism (\textit{\compd}, see Figure \ref{fig_model_framework}(d)) to tackle the challenge of region diversity in the urban area. Specifically, we optimize a gate function that adaptively moderates the master model with region-specific context to derive slave models for more accurate detection. The details of each component are introduced as follows.

\subsection{Stage One: Master Training Stage on the URG}
\label{rep_learn}
In our framework, we comprehensively use multi-modal region features and complex regions' relations to learn region representation for UV detection.
First, \compb performs local feature aggregation on URG with encoding both intra-modal and inter-modal context, to collectively enhance the region representation learning from both POI and satellite image modality. Then, \compc clusters the semantically similar regions in the urban area based on the aggregated features from \compb, and forms a hierarchical structure of the urban area. This structure enables message propagation between regions and clusters, to capture the long-range correlation among similar regions and share the global semantic context, which alleviates the label scarcity problem in UV detection. Finally, we train a master model based on \compb and \compc for UV detection, and associate each cluster with a pseudo label indicating whether the cluster contains known UVs, which can further enrich the inside regions' contextual information.

\subsubsection{\textbf{Mutual-Attentive Graph Aggregation}}
Given that multi-modal features can more comprehensively describe regions from different perspectives, the basic idea of \compb is to take advantage of mutual enhanced information across modalities for multi-modal fusion. Thus, in addition to the intra-modal features aggregation from the neighborhood that typical GNNs perform, our \compb updates the features of each modality with encoding the inter-modal context on URG to fuse and enhance the multi-modal region representation.
 
First of all, we generate the intra-modal context for each modality from neighboring regions through a self-attention mechanism, since the influence made by different regions around varies non-linearly. Specifically, taking the POI features $\bm{x}^P_i$ as an example, the attention score between regions is computed as:
\begin{equation}
    c^{P \leftarrow P}_{ij} = \sigma(\, \bm{a}_{P \leftarrow P}^T [\bm{W}_P \, \bm{x}^P_i \oplus \bm{W}_P \, \bm{x}^P_j] \,),
\label{eq:1}
\end{equation}
where $\bm{x}^P_j$ denotes the POI features of region $v_j$ adjacent to region $v_i$ (i.e. $j \in \mathcal{N}_i$ where $\mathcal{N}_i$ is the neighborhood of region $v_i$ on $\mathcal{G}$),
$\bm{W}_P$ is a trainable transformation matrix, 
$\oplus$ denotes the concatenate operation, $\bm{a}_{P \leftarrow P}$ is a trainable weight vector and $\sigma$ denotes a non-linear activation function (here it is LeakyReLu).
Then the representation of POI features with intra-modal context can be obtained through aggregating POI features from its neighbors, according to the contextual coefficient $\alpha^{P \leftarrow P}_{ij}$ normalized by \textit{Softmax} function:
\begin{align}
    \widehat{\bm{x}}^{P \leftarrow P}_i &= \sigma(\sum_{j \in \mathcal{N}_i} \!\! \alpha^{P \leftarrow P}_{ij} \bm{W}_P \bm{x}^P_j), \\
    \alpha^{P \leftarrow P}_{ij} &= \frac{exp(c^{P \leftarrow P}_{ij})}{\sum_{k \in \mathcal{N}_i} exp(c^{P \leftarrow P}_{ik})}.
\label{eq:2}
\end{align}
Similarly, we generate the representation vector $\widehat{\bm{x}}^{I\leftarrow I}_i$ of image features with the intra-modal context in the same way:
\begin{equation}
    \widehat{\bm{x}}^{I\leftarrow I}_i = \sigma(\sum_{j \in \mathcal{N}_i} \alpha^{I \leftarrow I}_{ij} \bm{W}_I \bm{x}^I_j),
\label{eq:3}
\end{equation}
where $\bm{x}^I_j$ denotes the image features of regions adjacent to region $v_i$, and we have the same parameters $\bm{W}_I$ and $\bm{a}_{I \leftarrow I}$ corresponding to $\bm{W}_P$ and $\bm{a}_{P \leftarrow P}$.
Thus, the local dependencies are captured for each modality. 

To achieve the multi-modal fusion, we further adopt a cross-modal graph attention layer to summarize the contextual information from another modality and update the current representation vector.
Formally, taking the POI features of region $v_i$ as an example, we compute another attention score across modalities by:
\begin{equation}
    c^{P \leftarrow I}_{ij} = \sigma(\, \bm{a}_{P \leftarrow I}^T [\bm{W}^{'}_P \, \bm{x}^P_i \oplus \bm{W}^{'}_I \, \bm{x}^I_j] \,).
\label{eq:4}
\end{equation}
In this procedure, \compb gathers visual context from adjacent regions to the POI representation of the current region, where $\bm{W}^{'}_P$ and $\bm{W}^{'}_I$ denote another set of parametrized transformation matrices for POI features and image features, respectively. And as before, $\bm{a}_{P \leftarrow I}$ is the weight vector used for generating the cross-modal attention score, which implies how important is the context extracted from image features of $v_j$ to the POI representation learning of region $v_i$. Based on the score, the inter-modal context for POI representation is represented as:
\begin{align}
    \widehat{\bm{x}}^{P \leftarrow I}_i &= \sigma(\sum_{j \in \mathcal{N}_i} \alpha^{P \leftarrow I}_{ij} \bm{W}^{'}_I \bm{x}^I_j), \\
    \alpha^{P \leftarrow I}_{ij} &= \frac{exp(c^{P \leftarrow I}_{ij})}{\sum_{k \in \mathcal{N}_i} exp(c^{P \leftarrow I}_{ik})}.
\label{eq:5}
\end{align}
Then we incorporate the inter-modal context and update the POI representation through an aggregation function, (which can be concatenation, summation and attention mechanism):
\begin{equation}
    \label{eq:6}
    \widehat{\bm{x}}^P_i = AGG \, (\widehat{\bm{x}}^{P \leftarrow P}_i, \widehat{\bm{x}}^{P \leftarrow I}_i).
\end{equation}
The updated local representation of image features is computed in the same way: $ \widehat{\bm{x}}^I_i = AGG \, (\widehat{\bm{x}}^{I \leftarrow I}_i, \widehat{\bm{x}}^{I \leftarrow P}_i)$.
To summarize, \compb integrates the inter-modal contextual information into feature aggregation process and enhances the representation of each modality. 
Note that only one layer aggregation procedure is presented above for simplicity, but in practice we can stack more layers to exploit richer contextual information cross modalities on URG for modal fusion.
Subsequently, the enriched multi-modal representation of regions can be obtained by $\widehat{\bm{x}}_i = \widehat{\bm{x}}^P_i \oplus \widehat{\bm{x}}^I_i$.

\begin{figure}[t]
\centering
\includegraphics[width=0.65\columnwidth]{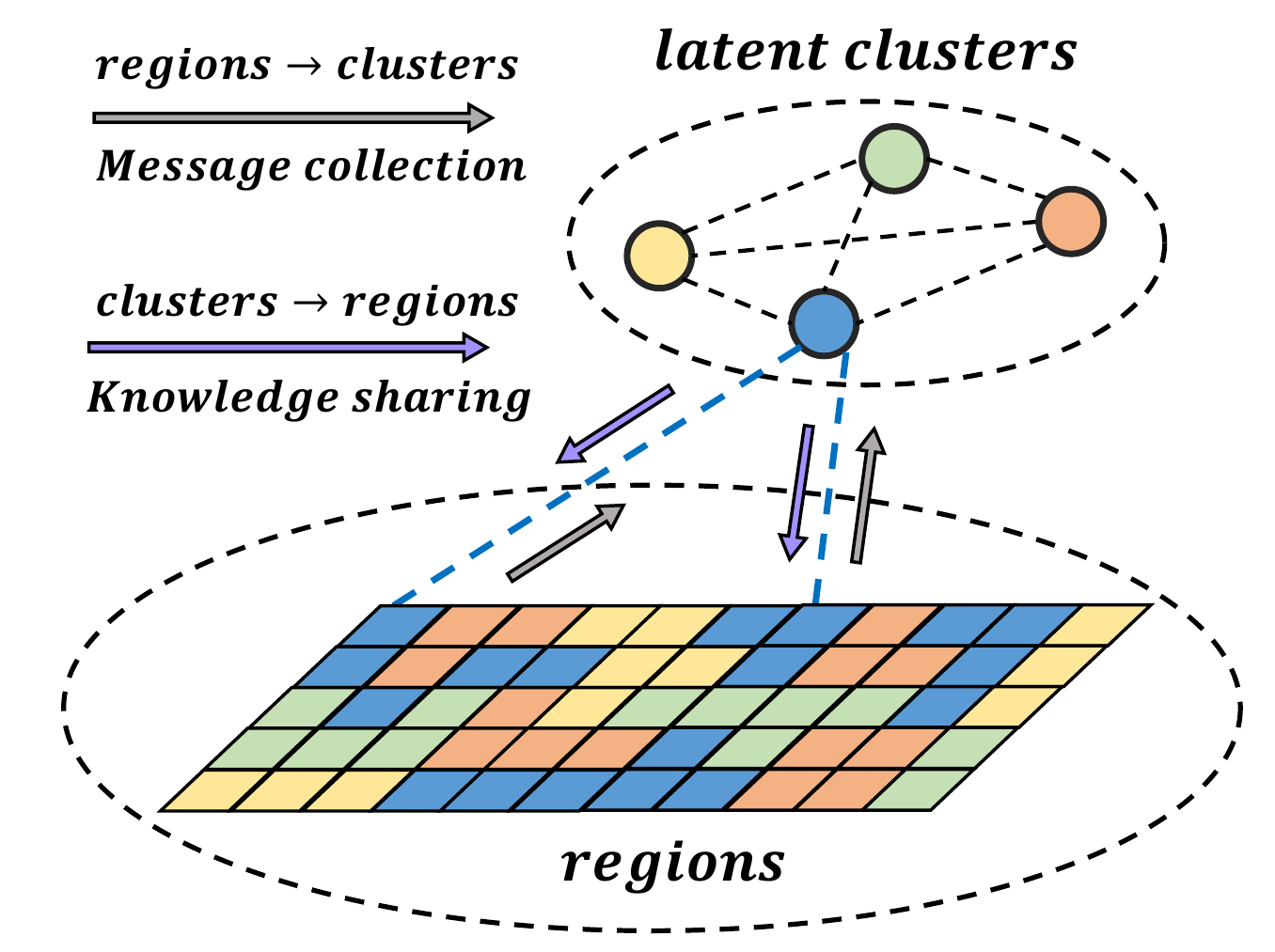}
\vspace{-1mm}
\caption{Illustration of the hierarchical structure over the URG. The region grids in the same color belong to the same latent cluster.}
\label{fig_hierarchical}
\vspace{-3mm}
\end{figure}

\subsubsection{\textbf{Global Semantic Clustering}}
After the local representation learning through \compb, 
\compc organizes the urban area as a hierarchical structure to cluster the semantically similar regions according to the multi-modal representation, and enables distant UVs to interact with each other through this structure. In this way, \mymodel not only models the complex dependency among regions on the URG, but also utilizes such region correlation to form the global semantic context and alleviate the label scarcity problem for UV detection.

In general, we assume that there are $K$ latent nodes standing for $K$ clusters of regions showing different characteristics in the urban area. \compc assigns regions into $K$ latent clusters and learns the cluster representation by performing $regions \rightarrow clusters$ message collection based on the local representation of regions derived by \compb. Subsequently, the learned cluster representation is propagated back in $clusters \rightarrow regions$ direction, which provides sharing global context among distant but similar regions. An illustration is shown in Figure \ref{fig_hierarchical}.

Formally, we define an assignment matrix $\bm{B} \in \mathbb{R}^{N \times K}$ to model the regions' membership to the $K$ clusters, where $\bm{B}_{ij}$ represents the probability of region-$i$ belonging to cluster-$j \in \{1,...,K\}$ and $\sum_{1 \le j \le K} \bm{B}_{ij} =1$. For region-$i$, after generating a local representation vector by \compb layers $\widetilde{\bm{x}}_i = \compb(\bm{x}^P_i, \bm{x}^I_i)$, we then apply a linear transformation followed by row-wise \textit{Softmax} function to compute the assignment matrix:
\begin{equation}
\label{eq:7}
    \bm{B} = Softmax(\bm{W}_B \, \widetilde{\bm{x}}_i),
\end{equation}
where $\bm{W}_B$ denotes the trainable weight of linear transformation. This assignment matrix serves as an information transmission channel between regions and clusters. \zhou{We further derive a binarized assignment matrix $\widetilde{\bm{B}}$ whose row is a one-hot vector with one at the position of maximal probability in the corresponding row of $\bm{B}$, i.e. $\widetilde{\bm{B}}_{ij}=1, j=argmax_k \bm{B}_{ik}$.} Then, the representation vectors of latent clusters are initialized by the weighted summation of regions' local representations according to this assignment matrix:
\begin{equation}
\label{eq:8}
    \bm{h}_j = \sum_{1 \le i \le N} \widetilde{\bm{B}}_{ij} \widetilde{\bm{x}}_i ,
\end{equation}
where $\bm{h}_j$ denotes the representation of cluster-$j$.
This binarization of the assignment matrix restricts each region to only one most likely cluster and avoids cluster representation being dominated by a large number of regions with very low membership probabilities.

After the above $regions \rightarrow clusters$ projection, the latent clusters segment the region set into $K$ groups, and their representation globally summarizes the similar semantic information of regions inside. Then, to capture the relation among clusters, we treat them as nodes in a complete graph with learnable edge weights, and reason their relevancy \hide{in our detection problem }by adopting graph convolution \cite{kipf2016semi}:
\begin{equation}
\label{eq:9}
    \bm{h}^{'}_{i} = \sigma(\sum_{1 \le j \le K} e_{ij} \, \bm{W}_{h} \, \bm{h}_j),
\end{equation}
where $\bm{h}^{'}_i$ denotes the updated representation of cluster-$i$, $\bm{W}_h$ denotes the shared transformation matrix and $e_{ij}$ is the edge weight corresponding to the influence of cluster-$j$ to cluster-$i$, which is trained together with $\bm{W}_h$. 
Once $\bm{h}^{'}_i$ is obtained, we reuse the assignment matrix and perform a $clusters \rightarrow regions$ reverse knowledge sharing to enhance region representation with this global context.
Rather than the binary assignment matrix $\widetilde{\bm{B}}$, we use the original soft assignment matrix $\bm{B}$ in this procedure since less-correlated clusters can still auxiliarily enrich the region representation. The reverse knowledge sharing is expressed as:
\begin{equation}
\label{eq:10}
    \widetilde{\bm{x}}^{g}_{i} = \sigma(\sum_{1 \le j \le K} \bm{B}_{ij} \, \bm{W}_{r} \, \bm{h}^{'}_j),
\end{equation}
where $\widetilde{\bm{x}}^{g}_{i}$ denotes the global-aware region representation reversed from latent clusters, $\bm{W}_r$ denotes the weights to be learned. The enhanced region representation is then derived by combining the local and global representation through an aggregation function:
\begin{equation}
\label{eq:11}
\widetilde{\bm{x}}^{'}_i = AGG(\widehat{\bm{x}}_i, \widetilde{\bm{x}}^{g}_i).
\end{equation}
In this way, global semantic information can be effectively shared across regions. More importantly, the undiscovered UVs are more likely to interact and exchange information with the limited known UVs through the global-aware part $\widetilde{\bm{x}}^{g}_i$, which alleviates the label scarcity problem in UV detection. 

\subsubsection{\textbf{Training the Master Model}} \label{stage1}
There are two goals in the master training stage. First, we can optimize \compb and \compc by training the master model for UV detection across all the regions in an end-to-end manner. After that, the region representation and hierarchical graph structure (i.e. the membership of regions to the latent clusters) can be learned.
Second, we associate each cluster with a pseudo label by performing a $regions \rightarrow clusters $ label collection based on the labels of regions inside this cluster. This pseudo label indicates whether the cluster contains known UVs. 

Formally, with the learned region representation $\widetilde{\bm{x}}^{'}_i$ described in (\ref{eq:11}), the master model is defined as the hierarchical graph neural network and a following classifier taking $\widetilde{\bm{x}}^{'}_i$ as input to identify whether region-$i$ is an UV:
\begin{equation}
\label{eq:12}
    \mathcal{M}(\widetilde{\bm{x}}^{'}_i, \Phi_{m}) \rightarrow y_i, \, y_i \in \{0,1\},
\end{equation}
where $\mathcal{M}(\cdot, \Phi_{m})$ denotes the classifier in the master model with parameters $\Phi_{m}$, which is a 2-layer Multi-Layer Perceptron (MLP) in this paper. 
We use binary cross entropy (BCE) to define the detection loss in this binary classification task:
\begin{equation}
\small
\label{eq:13}
    \mathcal{L}_c = \!\!\! \sum_{v_i \in V^L} \!\! -y_i log \, \mathcal{M}(\widetilde{\bm{x}}^{'}_i, \Phi_{m}) - (1-y_i) log (1 - \mathcal{M}(\widetilde{\bm{x}}^{'}_i, \Phi_{m})),
\end{equation}
where $V^L$ denotes the labeled region set. 

After the training process, the membership of regions formed by assignment matrix $\widetilde{\bm{B}}$ is fixed. We initiatively transmit the region labels through $\widetilde{\bm{B}}$ in the direction of $regions \rightarrow clusters$, and derive pseudo labels for latent clusters.
Specifically, using $y^h_j$ to denote the binary pseudo label of cluster $j$, we set $y^h_j = 1$ if there exists at least one known UV inside, otherwise $y^h_j=0$. The rule of pseudo label generation can be expressed as:
\begin{equation}
\label{eq:14}
    y^h_j = \left \{
    \begin{array}{ll}
      1,\,\,\, \sum_{1 \le i \le N} \widetilde{\bm{B}}_{ij} \, y_i > 0, \\
      0,\,\,\, \sum_{1 \le i \le N} \widetilde{\bm{B}}_{ij} \, y_i =  0. \\
        \end{array}
      \right.
\end{equation}
This pseudo label directly provides contextual information that the inside regions are closely correlated 
to known UVs, and the predictor should pay special attention to them.
The detailed process of the master training stage is in Algorithm \ref{alg-stage1}.

\begin{algorithm}
    \small
    \caption{Master Training Stage of \mymodel}
    \label{alg-stage1}
    \LinesNumbered
    \KwIn{URG $\mathcal{G}(\mathcal{V},\mathcal{E},\bm{A},\bm{X})$, number of latent clusters $K$}
    \KwOut{Trained master model with parameter set: $\bm{\theta}_1=\{\bm{W}_{\{P,I,B,h,r\}},$ 
    $\bm{W}^{'}_{\{P,I\}},$ 
    $\bm{a}^{\{P,I\}\leftarrow \{P,I\}},$
    $\Phi_{m}\}$}
    Randomly initialize the parameter set $\bm{\theta}_1$;\\
    \For{iteration = 1,2,3, ...}{
    Get multi-modal region representation by \eqref{eq:1}-\eqref{eq:6};\\
    Get assignment matrix $\bm{B}$ by \eqref{eq:7};\\
    Get cluster representation $\bm{h}^{'}_{i}$ by \eqref{eq:8}-\eqref{eq:9};\\
    Get region representation $\widetilde{\bm{x}}^{'}_i$ by \eqref{eq:10}-\eqref{eq:11};\\
    Get UV prediction $\mathcal{M}(\widetilde{\bm{x}}^{'}_i)$ with master model by \eqref{eq:12};\\
    Get the UV detection loss $\mathcal{L}_c$ by \eqref{eq:13};\\
    Update parameters $\bm{\theta_1}$ according to the gradient of $\mathcal{L}_c$;
    }
    Derive pseudo label of latent clusters $y^{h}_{j}$ by \eqref{eq:14};\\
    \Return{$\bm{\theta_1}, \bm{B}, y^{h}_{j}$}
\end{algorithm}

\subsection{Stage Two: Slave Adaptive Training Stage}\label{stage-2}
\label{gate}

In the second stage, the main objective is to train a gate function (see Figure \ref{fig_model_framework}(d)) which can
encode the contextual information to derive region-wise slave \xcx{models}. 
The classifier in the master model is also fine-tuned in this stage.  Here we propose a contextual master-slave gating mechanism (\compd), which uses the gate function to moderate the master model to derive slave models conditioned on regions' context vector.  
\xcxre{Note that there is a previous study applying a gating mechanism \cite{yao2019revisiting} for urban applications, but it is used to control the weight of information propagated between regions when using CNNs to capture local spatial dependency for traffic prediction. Thus, our \compd designed to derive slave models is different from it.}
After the master training stage, we can build a context vector for each region. For each latent cluster, we first estimate its possibility to include UVs, and then form the context vector for each region using this UV inclusion possibility (according to the soft assignment matrix for regions and clusters). The basic idea of building such a context vector is that if an unlabeled region is clustered together with known UVs, it should have a higher probability to be an UV. 
Thus, in the slave adaptive training stage, we first estimate the UV inclusion probability for each cluster based on the pseudo label, and then generate the context vector for each region through a $clusters \rightarrow regions$ probability transmission.

In general, the context vector for each region is formed by predicting the pseudo label for each cluster. In an urban area, only a limited number of UVs are known. Thus, there may be only a few clusters associated pseudo label as $1$ while the majority as $0$. However, it probably misguides the model if we simply assume the majority of clusters have no UVs, because in fact these clusters possibly contain some undiscovered UVs just need to be detected. Therefore, rather than directly use the pseudo label, we additionally exert a pseudo label predictor to estimate the inclusion probability that a latent cluster contains UVs. Specifically, we use $\mathcal{M}^p(\cdot, \Phi_p)$ to denote the pseudo label predictor parametrized by $\Phi_p$, which takes cluster representation as input and predicts pseudo labels by:
\begin{equation}
\label{eq:15}
    \widehat{y}^h_j = \mathcal{M}^{p}(\bm{h}^{'}_j, \Phi_{p}) \rightarrow y^h_j, \, y^h_j \in \{0,1\},
\end{equation}
where $\widehat{y}^h_j \in (0, 1)$ denotes the output inclusion probability, which ought to be higher for clusters with known UVs inside ($y^h_j$ = 1) than that for the others. 
Note that the inclusion probability estimation is actually a positive-unlabeled (PU) learning problem where the clusters with no labeled UVs are actually with unknown labels. Thus, following the previous PU learning method \cite{wei2018positive}, we define a rank loss function to optimize the pseudo label predictor:
\begin{equation}
\label{eq:16}
    \mathcal{L}_p = \sum_{c_i \in \mathcal{C}_1} \sum_{c_j \in \mathcal{C}_0} (1 - (\widehat{y}^h_i - \widehat{y}^h_j))^2,
\end{equation}
where $\mathcal{C}_1$ and $\mathcal{C}_0$ denote the clusters with and without known UVs, respectively. Guided by $\mathcal{L}_p$, the pseudo label predictor $\mathcal{M}^{p}$ learns to estimate how likely a cluster contains UVs.
Given this inclusion probability, the gate function learns to form the region-specific context vector and moderate the master model to derive the slave model.
First, the context vector for each region is formed through performing a $clusters \rightarrow regions$ inclusion probability transmission depending on the region's membership to every cluster by:
\begin{equation}
\label{eq:17}
    \bm{q}_i = \sigma(\bm{W}_q (\bm{B}_{i, *} \circ \widehat{\bm{Y}}^h)), \,\,\, \widehat{\bm{Y}}^h = [\widehat{y}^h_1, \widehat{y}^h_2, ..., \widehat{y}^h_K],
\end{equation}
where $\bm{q}_i$ denotes the region-specific context vector, $\bm{W}_q$ is the trainable weights, $\widehat{\bm{Y}}^h$ denotes the inclusion probability vector of all latent clusters, $\bm{B}_{i, *}$ denotes the row-$i$ of the assignment matrix and $\circ$ denotes the Hadamard product. 
This context reflects that the region-$i$ is correlated to some known or potential UVs with different probabilities.

Subsequently, we adaptively derive the slave model from the master model for each region by imposing this region-specific contextual information on its parameters with a gating \hide{function}\xcx{mechanism}. \hide{To be specific}\xcx{Specifically}, \hide{we}\xcx{the gate function} first generates an adaptive region-specific parameter filter from the context vector by:
\begin{equation}
\label{eq:18}
    \bm{\mathcal{F}}_i = \sigma(\bm{W}_f \, \bm{q}_{i}),
\end{equation}
where $\bm{\mathcal{F}}_i$ denotes the parameter filter of region-$i$, which has the same number of parameters with the classifier in the master model $\Phi_m$, $\bm{W}_f$ denotes the weight matrix linearly mapping context vector to the parameter space, and $\sigma$ here denotes the \textit{Sigmoid} activation function restricting the elements in filter into range $(0, 1)$. 
Then, a region-specific slave model with adaptive predictor $\mathcal{M}^i(\, \cdot \,, \Phi_m^i)$ can be derived by leveraging the filter to tailor the parameters of $\mathcal{M}(\, \cdot\, , \Phi_m)$ through the \compd mechanism, which can be formulated as:
\begin{equation}
\label{eq:19}
    \Phi_m^i = \mathcal{F}_i \circ \Phi_m,
\end{equation}
where $\Phi_m^i$ denotes the modified parameters of the new model customized for region-$i$.
With this region-specific slave \xcx{model}\hide{ predictor}, the final region-wise UV detection in this slave training stage is performed by:
\begin{equation}
\label{eq:20}
    \mathcal{M}^i (\widetilde{\bm{x}}^{'}_i, \Phi_{m}^i) \rightarrow y_i, \, y_i \in \{0,1\}.
\end{equation}
Correspondingly, the detection loss function is redefined as: 
\begin{equation}
\label{eq:21}
    \mathcal{L}_c^{'} = \!\!\! \sum_{v_i \in V^L} \!\! -y_i log \, \mathcal{M}^{i}(\widetilde{\bm{x}}^{'}_i, \Phi_{m}^{i}) - (1-y_i) log (1 - \mathcal{M}^{i}(\widetilde{\bm{x}}^{'}_i, \Phi_{m}^{i})).
\end{equation}
The optimization objective in the slave adaptive stage can be expressed by the weighted summation of pseudo label predicting loss and final UVs detecting loss controlled by a balancing hyper-parameter $\lambda$:
\vspace{-1mm}
\begin{equation}
\label{equ:22}
    \mathcal{L} = \mathcal{L}_c^{'} + \lambda \mathcal{L}_p.
\vspace{-1mm}
\end{equation}

The detailed training procedure of the slave adaptive stage is presented in Algorithm \ref{alg-stage2}.
\begin{algorithm}
    \small
    \caption{ Slave Adaptive Training Stage of \mymodel}
    \label{alg-stage2}
    \LinesNumbered
    \KwIn{URG $\mathcal{G}(\mathcal{V},\mathcal{E},\bm{A},\bm{X})$, number of latent clusters $K$, balancing hyper-parameter $\lambda$, trained master model with parameter set $\bm{\theta_1}$, assignment matrix $\bm{B}$, pseudo label of latent clusters $y^{h}_{j}$, gate function with $\bm{W}_{\{q,f\}}$}
    \KwOut{Trained \mymodel with parameter set: 
    $\bm{\theta_2} \!=\! \bm{\theta_1} \! \cup \! \{\bm{W}_{\{q,f\}}\!, \! \Phi_{p}\}$}
    Initialize the parameter set $\bm{\theta_1}$ with trained master model;\\
    Randomly initialize other parameters of $\bm{\theta_2} \setminus \bm{\theta_1}$;\\
    \For{iteration = 1,2,3, ...}{
    Get multi-modal region representation by \eqref{eq:1}-\eqref{eq:6};\\
    Get cluster representation $\bm{h}^{'}_{i}$ by \eqref{eq:8}-\eqref{eq:9};\\
    Get region representation $\widetilde{\bm{x}}^{'}_i$ by \eqref{eq:10}-\eqref{eq:11};\\
    Estimate inclusion probability $\widehat{y}^{h}_{j}$ by \eqref{eq:15};\\
    Get pseudo label prediction loss $\mathcal{L}_p$ by \eqref{eq:16};\\
    Get region-specific parameter filter $\mathcal{F}_i$ by \eqref{eq:17}-\eqref{eq:18};\\
    Get adaptive slave model with $\mathcal{M}(\, \cdot\, , \Phi_m)$ by \eqref{eq:19};\\
    Get final UV prediction $\mathcal{M}^{i}(\widetilde{\bm{x}}^{'}_i)$ by \eqref{eq:20};\\
    Get the updated UV detection loss $\mathcal{L}^{'}_c$ by \eqref{eq:21};\\
    Update parameters $\bm{\theta_2}$ according to the gradient of $\mathcal{L}$;
    }
    \Return{$\bm{\theta_2}$}
\end{algorithm}

\subsection{Urban Village Detection}
After the two-stage training, our \mymodel can make urban village detection across the city as follows. 
Given an unlabeled region on the URG, we first compute its membership to different clusters and produce the region-specific context vector based on this membership and clusters' inclusion probability. Then, the parameter filter can be further generated to gate the master model and derive the corresponding slave model. \xcx{At last, we feed the raw features of this region into the slave model to output the probability of being UV.}

\xcxre{
\subsection{Complexity Analysis}
Finally, we analyze the 
\zhoure{time} complexity of our \mymodel. Note that the processing steps in the master training stage are included in the slave adaptive stage, thus we analyze the computational \zhoure{time} cost of each component (\compb, \compc and \compd) per iteration in the slave stage as the overall complexity of \mymodel.
Specifically, feeding the URG as input, the complexity of \compb is:
\begin{equation}
\label{eq:23}
    \mathcal{T}_{\compb} = \mathcal{O}(|\mathcal{V}|d^{2} + |\mathcal{E}|d),
\end{equation}
where $|\mathcal{V}|$ and $|\mathcal{E}|$ denote the size of the node set and edge set of URG, $d$ denotes the dimension of region features. $\mathcal{O}(|\mathcal{V}|d^{2})$ is the cost of feature transformation, and $\mathcal{O}(|\mathcal{E}|d)$ corresponds to the complexity of attention score computation and feature aggregation. 
The complexity of \compc is:
\begin{equation}
\label{eq:24}
    \mathcal{T}_{\compc} = \mathcal{O}(|\mathcal{V}|Kd + Kd^2 + K^2d),
\end{equation}
where $K$ is the number of latent semantic clusters. $\mathcal{O}(|\mathcal{V}|Kd)$ represents the complexity of assigning all regions into $K$ clusters to get cluster representation, as well as obtaining global-aware region representation through reverse knowledge sharing from clusters. And $\mathcal{O}(Kd^2 + K^2d)$ denotes the cost of graph convolution operation among latent clusters. 
As for the \compd mechanism, we compute its complexity as:
\begin{equation}
\label{eq:25}
    \zhoure{\mathcal{T}_{\compd} = \mathcal{O}(Kd + |\mathcal{V}|K + |\mathcal{V}|Kd + |\mathcal{V}|d|\mathcal{F}_i| )},
\end{equation}
where $\mathcal{O}(Kd)$ is to estimate the UV inclusion probability of each cluster, and the region-wise context is formed by this inclusion probability vector with complexity of $\mathcal{O}(|\mathcal{V}|K + |\mathcal{V}|Kd)$. Then, denoting the parameter size of final region-specific predictor as $|\mathcal{F}_i|$, the computational cost of generating parameter filter and deriving slave model is \zhoure{ $\mathcal{O}(|\mathcal{V}|d|\mathcal{F}_i| + |\mathcal{V}||\mathcal{F}_i|)=\mathcal{O}(|\mathcal{V}|d|\mathcal{F}_i|)$ (since $d\geq 1$).} In our work, the parameter size of the predictor can be represented as \zhoure{$|\mathcal{F}_i|=\mathcal{O}(d^2)$}. 
Overall, the total complexity of \mymodel is the combination of $\mathcal{T}_{\compb}$, $\mathcal{T}_{\compc}$ and $\mathcal{T}_{\compd}$:
\vspace{-1mm}
\begin{equation}
\label{eq:26}
    \mathcal{T}_{\mymodel} = \mathcal{O}(|\mathcal{V}|d^3 + |\mathcal{V}|Kd + |\mathcal{E}|d + Kd^2 + K^2d).
\end{equation}
}

\vspace{-4.5mm}
\section{EXPERIMENTS}
In this section, we conduct experiments in three cities in China to demonstrate the effectiveness of our framework.

\subsection{Experimental Setup}
\begin{table}[t]
	\caption{Statistics of three real-world datasets.}
	\vspace{-2mm}
	\label{exp-dataset}
    \centering
    \resizebox{0.45\textwidth}{!}{
	\begin{tabular}{c|c|c|p{1cm}<{\centering}|c}
		\toprule
		 & \# Regions & \# Edges & \# UVs & \# Non-UVs \\
		\cline{1-5}
		\rule{0pt}{10pt}
		Shenzhen & 93,600 & 3,624,676 & 295 & 6,867\\
		\cline{1-5}
		\rule{0pt}{10pt}
		Fuzhou & 59,872 & 1,589,198 & 276 & 3,685\\
		\cline{1-5}
		\rule{0pt}{10pt}
		Beijing & 354,316 & 19,086,524 & 204 & 10,861\\
		\bottomrule
	\end{tabular}
	}
	\vspace{-5mm}
\end{table}
\textbf{Data collection.} We evaluate the performance of the proposed framework \mymodel on three real-world datasets with POI data, satellite image data, road network data, and ground-truth binary label data in Shenzhen, Fuzhou, and Beijing. For each city, the POI basic property data and satellite image data used for region features construction are also collected by Baidu Maps in June 2020, while the road network data is collected by \cite{karduni2016protocol}.
Besides, the ground-truth UV and non-UV regions for the three real-world datasets in our work are collected through exhaustive manual crowdsourcing in June 2020.
\xcxre{More information about how to collect the ground-truth data is introduced in Appendix I-C.}

\textbf{Datasets construction.} Upon these collected data, the three real-world datasets are constructed as follows. We divide the city into $128m \times 128m$ region grids, and our datasets include the regions in the main urban area. 
In this experiment, the main urban area is defined as region grids selected by a centered rectangle frame covering $90\%$ POIs in the city. After data cleaning and coordinate alignment, we obtained the three real-world datasets whose statistical information are summarized in Table \ref{exp-dataset}, including the total number of region grids, edges, as well as the labeled samples of UVs and non-UVs. 

To achieve stable experiment results, we selected the optimal hyper-parameters for each model based on 3-fold nested cross-validation.
Specifically, we first equally split the dataset into three folds, where each fold will serve as test data in turn for performance evaluation, then the rest two folds are used for model training and parameters selection with another 2-fold cross-validation. We report the average results of three rounds in each experiment. 

Moreover, to avoid potential information leakage in real-life applications, we use a coarse-grained partition strategy to split the dataset for cross-validation.
Notably, in practice, the unlabeled region grids are usually distributed in patches (e.g. a residential area composed of a cluster of grids), and should not be mixed with labeled grids. Following the previous splitting method \cite{jean2019tile2vec}, we simulate this practical scenario by treating every $10 \times 10$ grids as a block and then performing data split on this coarse-grained block level. In this way, the labeled and unlabeled grids will not be mixed together.

\textbf{Implementations.}
For all comparing approaches in our experiments, we construct a 64-dimension POI features vector and generate a 4096-dimension image semantic features vector from the satellite image for each region as model inputs. If without specification, we use Adam optimizer with an initial learning rate of 0.0001, and the hidden size is set to 64.

For our \mymodel, we adopt an exponential decay strategy whose decay rate is set to $0.1\%$ per epoch in the optimization process. For \compb, the head number of multi-head attention is set to 2 for Shenzhen and Fuzhou, and 1 for Beijing.
We first apply a linear transformation to reduce the dimension of image features to 128, and stack two \compb layers to learn the multi-modal representation of regions with the aggregation function instantiated by the attention mechanism. For \compc, we set the number of latent clusters to 50, 500, 500 for Shenzhen, Fuzhou and Beijing. Note that when applying the softmax function to compute the assignment matrix, we introduce a temperature parameter $\tau$ \cite{lin2021graph} for constraining the membership probability to different clusters, where we set $\tau$ to 0.1, 0.01 and 0.1 for Shenzhen, Fuzhou and Beijing. We use one graph convolution layer to reason the correlation among clusters. The learned global-aware representation and local representation from \compb are aggregated by summation in Shenzhen, Fuzhou and concatenation in Beijing. For \compd, the pseudo label predictor is a simple LR classifier and the balancing weight $\lambda$ in the slave adaptive stage is set to $0.01, 1.0$ and $0.001$ for Shenzhen, Fuzhou and Beijing.

\renewcommand\arraystretch{1.2}
\begin{table*}[t]
    \small
	\caption{\xcxre{Detection Performance Comparison in Terms of Precision, Recall, and F1-score in three cities. The average and standard deviation (shown in brackets) results are reported across five random runs.}}
	\label{exp-main-3&5-meanstd}
	\vspace{-3mm}
    \centering
    \resizebox{0.92\textwidth}{!}{
	\begin{tabular}{c|c|c|c|c|c|c|c|c}
		\toprule
		\multicolumn{2}{c|}{\multirow{2}{*}{}} &
		\multirow{2}{*}[-3pt]{AUC} &
		\multicolumn{3}{c|}{$p=3$} & \multicolumn{3}{c}{$p=5$} \\
		\cline{4-9}
	    \multicolumn{2}{c|}{} &
	    \rule{0pt}{12pt}
	    & Recall & Precision & F1-score & Recall & Precision & F1-score \\
        \midrule
        \multirow{7}{*}{Fuzhou}
        & MLP & 0.837 (.001) & 0.145 (.007) & 0.376 (.013) & 0.208 (.009) & 0.250 (.010) & 0.371 (.011) & 0.295 (.010) \\
        & GCN & 0.831 (.003) & 0.149 (.006) & 0.365 (.016) & 0.209 (.009) & 0.220 (.010) & 0.325 (.013) & 0.259 (.011) \\
        & GAT & 0.850 (.010) & 0.160 (.017) & 0.391 (.043) & 0.224 (.024) & 0.244 (.010) & 0.352 (.017) & 0.284 (.012) \\
        & MMRE & 0.836 (.005) & 0.160 (.005) & 0.398 (.014) & 0.226 (.007) & 0.254 (.008) & 0.371 (.013) & 0.298 (.010) \\
        & UVLens & 0.854 (.004) & 0.161 (.011) & 0.389 (.025) & 0.225 (.015) & 0.256 (.009) & 0.368 (.012) & 0.298 (.010) \\
        & \xcxre{MUVFCN} & 0.846 (.004) & \underline{0.173 (.008)} & \underline{0.421 (.015)} & \underline{0.242 (.010)} & \underline{0.273 (.003)} & \underline{0.390 (.006)} & \underline{0.317 (.004)} \\
        & ImGAGN & \underline{0.865 (.001)} & 0.120 (.003) & 0.297 (.007) & 0.169 (.004) & 0.210 (.003) & 0.311 (.004) & 0.248 (.003) \\
        & \mymodel & \textbf{0.870 (.001)} & \textbf{0.181 (.003)} & \textbf{0.437 (.007)} & \textbf{0.253 (.004)} & \textbf{0.276 (.000)} & \textbf{0.391 (.001)} & \textbf{0.319 (.001)} \\
        \midrule
		\multirow{7}{*}{Shenzhen}
        & MLP & 0.691 (.001) & 0.090 (.003) & 0.123 (.004) & 0.103 (.003) & 0.149 (.002) & 0.122 (.002) & 0.134 (.002) \\
        & GCN & 0.598 (.019) & 0.040 (.008) & 0.059 (.011) & 0.048 (.009) & 0.069 (.006) & 0.061 (.005) & 0.064 (.005) \\
        & GAT & 0.669 (.023) & 0.075 (.008) & 0.098 (.011) & 0.085 (.009) & 0.115 (.016) & 0.093 (.013) & 0.102 (.014) \\
        & MMRE & 0.690 (.007) & 0.087 (.003) & 0.119 (.004) & 0.100 (.003) & 0.136 (.004) & 0.113 (.003) & 0.123 (.003) \\
        & UVLens & 0.713 (.015) & 0.105 (.016) & 0.140 (.020) & 0.119 (.017) & \underline{0.170 (.020)} & \underline{0.135 (.015)} & \underline{0.150 (.017)} \\
        & \xcxre{MUVFCN} & \underline{0.719 (.010)} & \underline{0.107 (.009)} & \underline{0.141 (.010)} & \underline{0.121 (.009)} & 0.162 (.012) & 0.128 (.008) & 0.142 (.010) \\
        & ImGAGN & 0.636 (.028) & 0.063 (.005) & 0.087 (.008) & 0.073 (.007) & 0.103 (.010) & 0.085 (.008) & 0.093 (.009) \\
        & \mymodel & \textbf{0.762 (.000)} & \textbf{0.110 (.001)} & \textbf{0.148 (.002)} & \textbf{0.126 (.002)} & \textbf{0.172 (.003)} & \textbf{0.139 (.002)} & \textbf{0.153 (.002)} \\
        \midrule
        \multirow{7}{*}{Beijing}
        & MLP & 0.699 (.003) & 0.208 (.002) & 0.135 (.001) & 0.155 (.001) & 0.277 (.004) & 0.107 (.001) & 0.148 (.002) \\
        & GCN & 0.715 (.006) & 0.136 (.009) & 0.092 (.004) & 0.102 (.005) & 0.226 (.015) & 0.085 (.004) & 0.116 (.005) \\
        & GAT & \underline{0.782 (.008)} & 0.254 (.014) & 0.160 (.009) & 0.185 (.009) & \underline{0.383 (.020)} & \underline{0.140 (.008)} & \underline{0.194 (.011)} \\
        & MMRE & 0.691 (.011) & 0.198 (.007) & 0.130 (.003) & 0.149 (.004) & 0.263 (.010) & 0.102 (.004) & 0.141 (.006) \\
        & UVLens & 0.772 (.007) & \underline{0.289 (.018)} & \underline{0.176 (.007)} & \underline{0.206 (.009)} & 0.375 (.015) & 0.136 (.002) & 0.190 (.004) \\
        & \xcxre{MUVFCN} & 0.750 (.015) & 0.258 (.021) & 0.159 (.006) & 0.186 (.008) & 0.336 (.031) & 0.125 (.005) & 0.174 (.008) \\
        & ImGAGN & 0.698 (.011) & 0.145 (.010) & 0.068 (.009) & 0.086 (.009) & 0.189 (.016) & 0.058 (.009) & 0.084 (.011) \\
        & \mymodel & \textbf{0.821 (.000)} & \textbf{0.299 (.001)} & \textbf{0.191 (.001)} & \textbf{0.221 (.001)} & \textbf{0.400 (.002)} & \textbf{0.149 (.000)} & \textbf{0.207 (.000)} \\
        \bottomrule
	\end{tabular}
	}
	\vspace{-3mm}
\end{table*}
\vspace{-1mm}
\subsection{Baselines}
To evaluate the performance of \mymodel, we compare it with several comparative methods:
Multi-layer Perceptron (MLP), GNN models (GCN \cite{kipf2016semi} and GAT \cite{velivckovic2018graph}), and state-of-the-art methods for UV detection (UVLens \cite{chen2021uvlens} \xcxre{and MUVFCN \cite{mast2020mapping}}), urban region recognition (MMRE \cite{jenkins2019unsupervised}), and imbalance graph embedding (ImGAGN \cite{qu2021imgagn}). 
\xcxre{The detailed description and implementation of baselines are listed in Appendix I-A.}

\vspace{-1mm}
\subsection{Evaluation Metrics}
To quantitatively measure the urban village detecting performance of \mymodel and the comparing methods, we use \textit{Area Under Curve} ($AUC$), $Recall$, $Precision$, and $F1-score$ as evaluation metrics.
Note that in the real-life application, the UV detection model is expected to screen out a small portion of potential UV candidates for facilitating the city manager to further investigate these regions with acceptable labor costs. Thus, we define $Recall$ and $Precision$ of UV detection in a practical application setting, where the top-$p\%$ regions with the highest probability ranked by the detection model are treated as the predicted UVs in the urban area of interest. Then, we compare these predicted UVs with ground truth to  calculate $Recall$ and $Precision$. In our experiments, we set $p=3$ and $p=5$ to evaluate the performance of all methods.

\subsection{Performance Comparison}
\label{exp-overall}
We first evaluate the performance of \mymodel and baseline approaches in urban village detection on the three real-world datasets. 
As shown in Table \ref{exp-main-3&5-meanstd}, our framework achieves the best performance. Compared with the best baselines, \mymodel improves the AUC by $6.8\%, 0.6\%$ and $5.0\%$ in Shenzhen, Fuzhou and Beijing respectively, which indicates that our method can more effectively detect the potential UVs.

Moreover, we further have the following observations.
Though without outstanding detection performance, MLP can positively discover UVs, which verifies the effectiveness of our constructed POI features and image features. 
Compared to MLP, GAT achieves great improvements in most cases, which demonstrates that instead of investigating every region independently, taking into account their correlations certainly benefits UV detection. 
However, also belonging to GNNs, GCN shows relatively poor performance in our problem. A possible reason is that GCN treats all the neighboring regions equally without capturing their relations to the current region and considering their different importance, which should provide useful information for UV detection. 
Since the region embedding method MMRE tries to fuse the POI and image features while leaning the region representation, it broadly outperforms GCN, suggesting that it's reasonable to further consider inter-modal contexts for enhancing multi-modal region representation.

\xcxre{
The state-of-the-art UV detection approaches UVLens and MUVFCN are the two most competitive solutions, but they still perform worse than our framework, which can be partially attributed to two major reasons: 
(1) Treating each region individually, they cannot capture the complex correlations among regions in an urban area, which plays a critical role in UV detection;
(2) These two approaches train a deep convolutional neural network without considering the scarcity of labeled UVs, which may significantly impact their performance.}
For the imbalanced network embedding method ImGAGN, despite considering the scarcity of known UVs and applying data augmentation to generate fake nodes and edges, it still cannot perform well because the augmented data lose the original relation and structure among regions in the urban context.

Compared with these solutions, our \mymodel framework consistently performs the best in terms of all metrics in three cities, thanks to its following advantages: (1) we construct an URG to comprehensively characterize region features and model complex dependencies among regions; (2) it gives full play to complementary advantages of inter-modal contexts to enhance the multi-modal region representation; (3) it enables the interactions among distant but similar regions through a hierarchical structure, to make exhaustive use of the knowledge from limited known UVs and alleviate the labeled UV scarcity problem; (4) the contextual master-slave gating mechanism improves the adaptability to diverse urban regions without the sacrifice of the generality. Therefore, our framework can much more effectively solve the UV detection problem.
\xcxre{Table \ref{exp-main-3&5-meanstd} also shows the standard deviation ($SD$) of all metrics for error analysis. As we can see, the $SD$ of AUC for all baselines is relative small, while the $SD$ of other metrics like Recall, Precision and F1-score are slightly larger. But overall, the performance of \mymodel is better than all competitors.
}

\begin{figure}
  \centering
  \subfigure[Effect of model components]{
    \includegraphics[width=0.45\columnwidth]{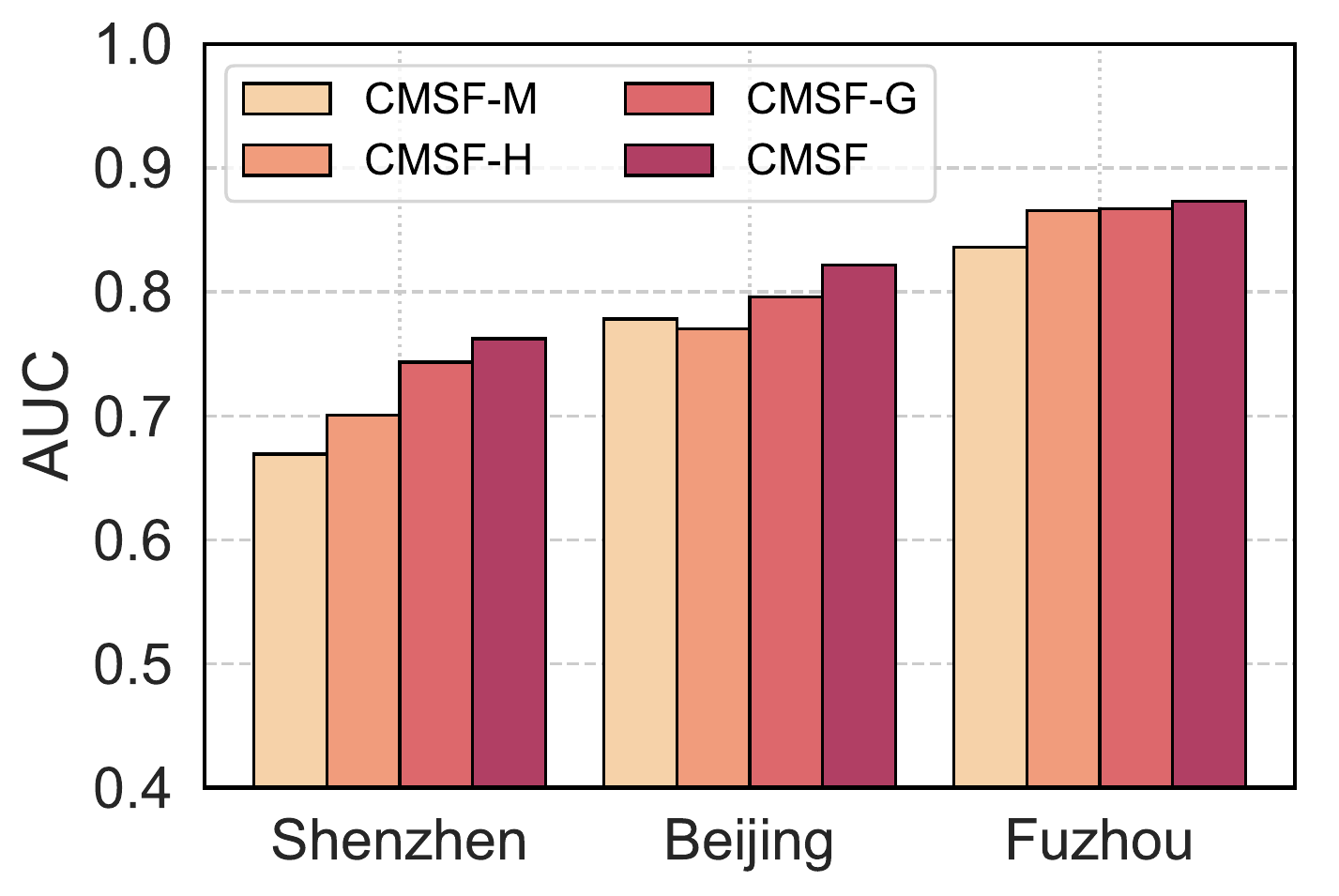}}
  \hspace{-2mm}
  \subfigure[\xcxre{Effect of multi-modal urban data}]{
    \includegraphics[width=0.45\columnwidth]{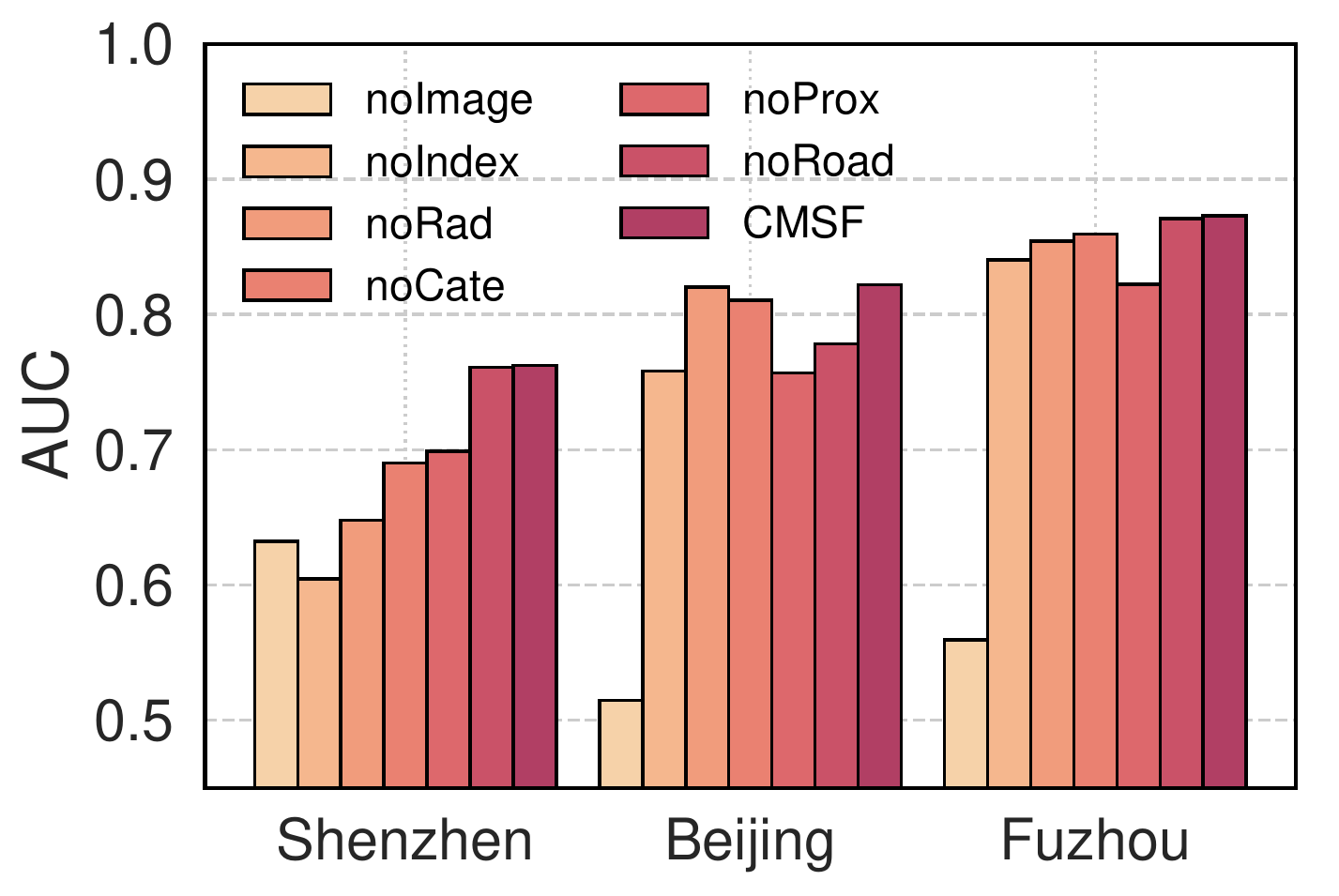}}
  \vspace{-3mm}
  \caption{\xcxre{Ablation studies of different factors in \mymodel.}}
  \label{fig-exp-ablation}
  \vspace{-4mm}
\end{figure}

\subsection{Ablation Study}
\label{exp-ablation}
\xcxre{
To verify the effectiveness of different factors in our proposed framework, we conduct the following two groups of ablation studies by comparing \mymodel with its variants on three datasets: (1) ablation of the designed components and (2) ablation of multi-modal urban data.}

\textit{\textbf{1) Effect of model components:}}
We first investigate the contributions of three components in our framework to UV detection by comparing \mymodel with its following variants:
\begin{itemize}[leftmargin=*]
    \item \textbf{\mymodel-M}. This variant uses vanilla GAT layers to replace \compb for learning region representation without taking into account the inter-modal context.
    \item \textbf{\mymodel-G}. This variant removes the \compd and omits the slave adaptive training stage, which directly uses the master model shared across all regions for the final prediction. 
    \item \textbf{\mymodel-H}. This variant removes the hierarchical structure including \compc and \compd. As a result, distant but similar regions are unable to interact with each other.
\end{itemize}
As shown in Figure \ref{fig-exp-ablation}(a), \mymodel outperforms all its variants, proving the significance of our special designs of contextual master-slave gating mechanism and the hierarchical graph neural network.
To be specific, \mymodel-M performs worse than \mymodel and other variants, since it independently aggregates the features of each modality, which indicates that the inter-modal context certainly benefits region representation learning and UV detection.
Besides, if we remove the contextual master-slave gating mechanism (\mymodel-G), the detection performance has a notable decline, suggesting the effectiveness of such a novel mechanism that balances the generality and specificity.
While the hierarchical structure is further removed (\mymodel-H), the performance gets worse. It indicates the necessity of interactions among distant but similar regions to share the global contextual information and alleviate the labeled UV scarcity problem in UV detection.

\xcxre{
\textit{\textbf{2) Effect of multi-modal urban data:}}
In addition to the methodology, we further analyze the contributions of multiple sources of urban data used to construct URG in our framework. In this experiment, we run \mymodel on the following changed URGs with different types of urban data removed.
\begin{itemize}[leftmargin=*]
    \item \textbf{noImage}. It removes the visual features from the satellite images, and regions are only characterized by POI features.
    \item \textbf{noCate}. Category distribution is not included. POI features only contain POI radius and index of basic living facility.
    \item \textbf{noRad}. POI radius is not included. POI features only contain category distribution and index of basic living facility.
    \item \textbf{noIndex}. Index of basic living facility is not included. POI features only contain category distribution and POI radius.
    \item \textbf{noRoad}. The edge set of urban region graph for this variant is only built by the spatial proximity defined by location.
    \item \textbf{noProx}. The edge set of urban region graph for this variant is only built by the connectivity on the road network.
\end{itemize}
Figure \ref{fig-exp-ablation}(b) presents the comparison results. We can have the following observations. Firstly, our CMSF beats the four variants which remove visual features (noImage) or one of the three types of POI features (noCate, noRad and noIndex). It proves the important role of satellite image features and our carefully designed POI features in profiling the region for UV detection. Secondly, another two variants that only consider the single region relation of spatial proximity (noRoad) or road connectivity (noProx) cannot perform well as CMSF, which indicates that modeling complex dependencies among regions from both spatial distance and road connectivity perspectives benefits more accurate UV detection.
}

\begin{figure}
  \centering
  \includegraphics[width=0.95\columnwidth]{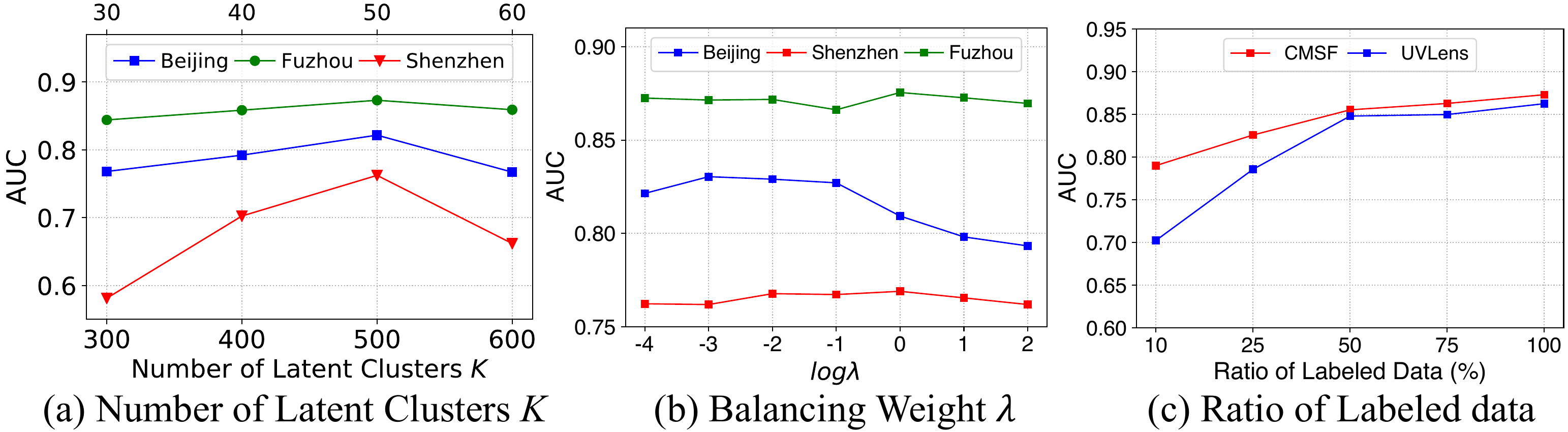}
  \vspace{-3mm}
  \caption{\xcxre{Parameter sensitivity of \mymodel. In (a), the bottom horizontal axis is for Fuzhou and Beijing datasets, while the upper one is for Shenzhen dataset.}}
\label{fig-exp-param-K}
\vspace{-4mm}
\end{figure}

\subsection{Parameters Analysis}
\label{exp-parameters}
We also investigate the hyper-parameter sensitivity, by evaluating the performance variation along with the change of each parameter while keeping other parameters fixed.

\textbf{Number of latent semantic clusters $K$.}
We first analyze the influence of the number of latent semantic clusters $K$. As depicted in Figure \ref{fig-exp-param-K}(a), with increasing $K$, \mymodel can model more complex and diverse latent semantic information in the urban context, leading to the rise of performance. However, if $K$ gets too large, there are not so many corresponding latent semantic groups in the urban area supporting the cluster representation learning, some superfluous clusters may result in more noise and undermine the performance instead.
Moreover, we observe that datasets in different cities prefer different $K$. A probable explanation is that the number of latent semantic groups is related to city area size, and the larger city (the area size of Beijing and Fuzhou is several times of the one of Shenzhen) may need a larger $K$ for complete modeling. 

\xcxre{
\textbf{Balancing weight $\lambda$.}
We next evaluate the sensitivity of the balancing weight $\lambda$. It can be observed in Figure \ref{fig-exp-param-K}(b) that the \hide{detection }performance arises first and then \zhoure{declines} 
when increasing $\lambda$, which indicates that applying the pseudo label prediction to regularize the region context with an appropriate weight to derive contextual slave models can improve the \hide{UV }detection performance. \zhoure{Whereas,} 
excessive focus on this objective \zhoure{(i.e. when $\lambda$ is relatively large)} can \hide{also }interfere the training process. 

\textbf{Ratio of labeled data.}
To verify the advantage of \mymodel to alleviate the scarcity problem of labeled UVs, we compare the performance of \mymodel and the most competitive baseline UVLens, with a different number of available labeled data for model training.
To be specific, we create four random masks that operate on the training set, to control the ratio of available labeled data to be 10\%, 25\%, 50\% and 75\%, and present the performance variation of these two methods trained on the four masked training sets. 
In Figure \ref{fig-exp-param-K}(c), we can observe that under different ratios of labeled data, \mymodel consistently outperforms the UVLens baseline. Moreover, with the change of labeling ratio, \mymodel presents a more stable performance variation and less degradation than UVLens when the labeled data becomes further scarce. These results further demonstrate the effectiveness of \mymodel to alleviate the scarcity of labeled UVs for real-world UV detection.}

\begin{table}[t]
\renewcommand\arraystretch{1.0}
    \small
	\caption{\xcxre{Efficiency comparison in Shenzhen and Fuzhou.}}
	\label{exp-time}
	\vspace{-3mm}
    \centering
    \resizebox{0.46\textwidth}{!}{
    \begin{tabular}{c|c|c|c|c|c}
		\toprule
		\multirow{2}{*}{} & 
		\multicolumn{2}{c|}{Training time(s)} & \multicolumn{2}{c|}{Inference time(s)} & 
		\xcxre{Model Size} \\
		\cline{2-5}
		\rule{0pt}{10pt}
		& Shenzhen & Fuzhou & Shenzhen & Fuzhou & \xcxre{(MBytes)} \\
		\midrule
        MLP & 0.075 & 0.032 & 0.037 & 0.012 & 1.048 \\
        GCN & 0.022 & 0.021 & 0.010 & 0.009 & 2.159 \\
        GAT & 0.053 & 0.040 & 0.026 & 0.022 & 2.369 \\
        MMRE & 240.4 & 116.7 & 0.002 & 0.002 & 3.981 \\
        UVLens & 0.369 & 0.443 & 0.194 & 0.189 & 450.1 \\
        \xcxre{MUVFCN} & 0.607 & 0.645 & 0.271 & 0.264 & 91.37 \\
        ImGAGN & 0.042 & 0.026 & 0.016 & 0.008 & 133.5 \\
        \mymodel & 0.187 & 0.342 & 0.112 & 0.062 & 7.433 \\
		\bottomrule
	\end{tabular}
	}
	\vspace{-4mm}
\end{table}

\subsection{Efficiency Comparison}
\xcxre{To comprehensively evaluate the efficiency, we compare all the baselines and our framework in terms of the training time and inference time, which stand for the offline training efficiency and deployed UV detection efficiency, as well as the parameter size that indicates the required space to apply the model. Specifically, we compute the average time of one epoch as the training time, while the inference time refers to the processing time for models obtaining the output probability from raw input. Here only the results in Shenzhen and Fuzhou are presented in Table \ref{exp-time} due to the limited space. For the parameter size, we only report the one in Fuzhou since the models for the three datasets have almost the same size.

We can observe that MLP, GCN, and GAT are most efficient in both time consumption and space requirement due to their simple structures, but it comes at the cost of unsatisfactory detection accuracy.
Rather, though UVLens and MUVFCN achieve better performance \zhoure{than MLP}, their \zhoure{operations} on high-dimensional image inputs inevitably result in a large parameter size and need much more time for intensive computation.
MMRE takes a lot of time in the training stage because its embedding model is partially optimized by an auxiliary task with time-costly negative sampling for each node on the graph. ImGAGN has a large model size, mainly due to the module that generates numerous links between the synthetic and minority nodes.
As for our \mymodel framework, we report the average time in the master training stage as training time, since it accounts for the most part of the whole training process, while the slave adaptive stage only needs very few iterations. 
Compared with UVLens and MUVFCN model, \mymodel not only achieves better performance, but also is much more efficient in both computational cost and model size. Therefore, our method has good efficiency in both time and space to achieve the best UV detection accuracy.
}

\subsection{Case Study}
\begin{figure}[t]
  \centering
  \includegraphics[width=0.9\columnwidth]{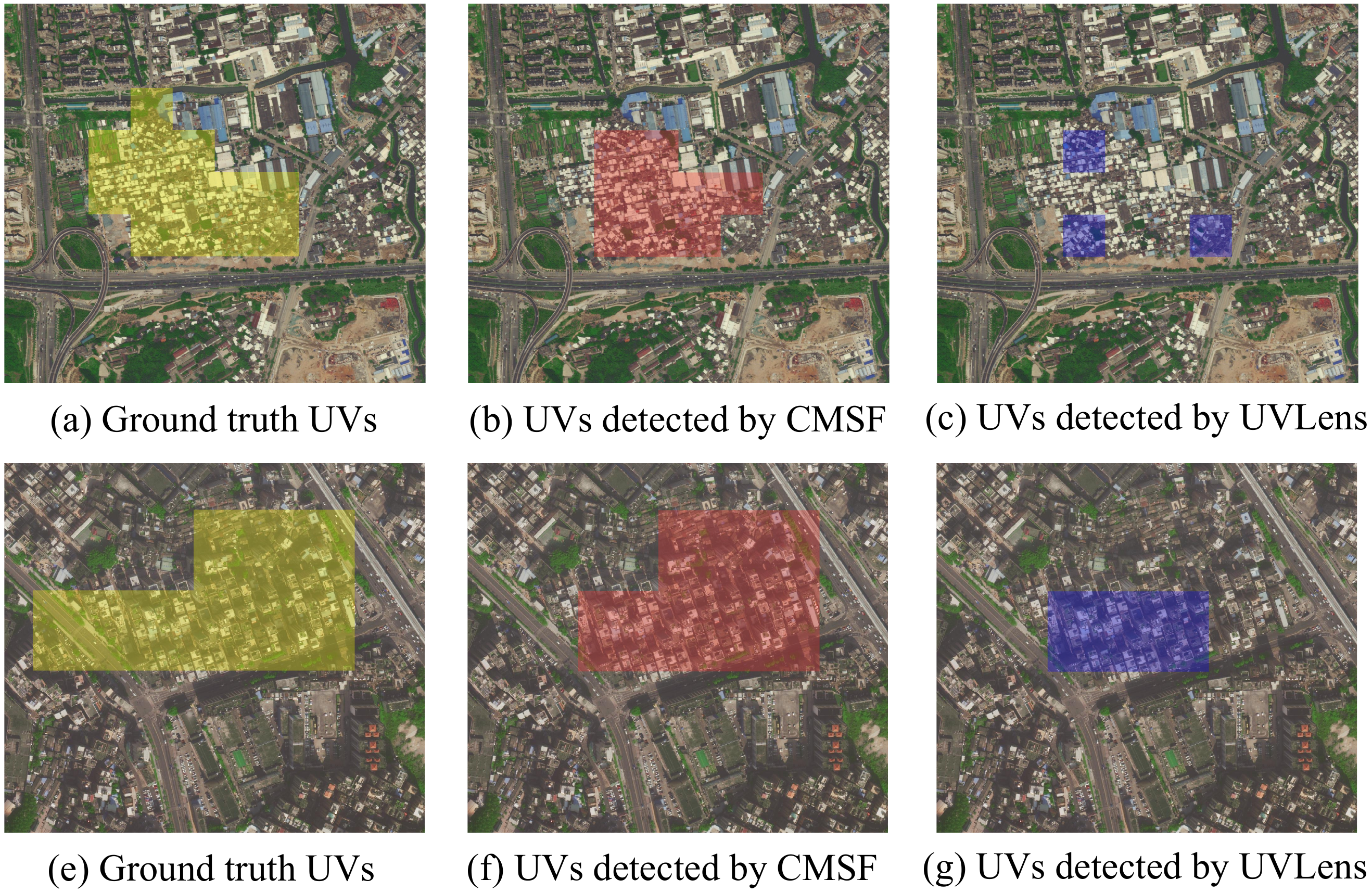}
  \vspace{-3mm}
  \caption{\xcxre{Case studies in Fuzhou (top row) and Shenzhen (bottom row).}}
  \label{fig-case}
  \vspace{-4mm}
\end{figure}

\xcxre{
Finally, we show cases in Fuzhou and Shenzhen in Figure \ref{fig-case}, to further demonstrate the advantage of \mymodel. 
Specifically, we apply the trained \mymodel and the state-of-the-art method UVLens to rank the regions in the labeled data based on their output probability, and then select the top 3\% ($p=3$) regions with the highest probability as detected UVs to compare with the ground truth labels. From the comparison shown in Figure \ref{fig-case}, we can observe that the regions detected by our \mymodel method (in red) evidently match better than those detected by UVLens (in blue) with the ground truth (in yellow), especially the surrounding UV areas of an apparent UV region. This is mainly because \mymodel can consider and utilize the dependencies among regions, which helps to effectively detect those highly correlated UVs together. 
}

\section{CONCLUSION}
In this paper, we investigate the urban village (UV) detection problem from the graph perspective. First, we construct an urban region graph (URG) incorporating multi-modal urban data. Then, we propose a Contextual Master-Slave Framework (\mymodel) over the URG to improve the performance of UV detection, which is trained in two stages. In the master training stage, we pre-train a hierarchical graph neural network as the master model to learn region representation and extract rich contextual information from the URG. In the slave adaptive stage, we devise a novel \compd mechanism to adaptively derive slave models for each region with region-specific contexts, which effectively balances the generality and specificity of our framework.
Extensive experimental results in three cities demonstrate the advantages of \mymodel to detect potential UVs.
In our future work, we plan to further investigate how to apply our framework to other urban applications.

\bibliographystyle{IEEEtran}
\bibliography{IEEEabrv, ref_abb}

\clearpage
\section*{Appendix}
\begin{table*}[t]
\renewcommand\arraystretch{1.2}
	\caption{Types of POI related to POI features construction.}
	\label{poi_type}
    \centering
	\begin{tabular}{p{2.83cm}<{\centering}|l}
		\toprule
		 Category Distribution &
		\begin{tabular}{l}
		\textbf{The category distribution features are calculated by the POI proportion of the following 23 categories:}\\
		Food Service, Hotel, Shopping Place, Life Service, Beauty Industry, Scenic Spot, Leisure and Entertainment, Sports\\
		and Fitness, Education, Cultural Media, Medicine, Auto Service, Transportation Facility, Financial Service, Real Estate,\\
		Company, Government Apparatus, Entrance and Exit, Topographical Object, Road, Railway, Greenland, Bus Route.
		\end{tabular}\\
        \midrule
		POI Radius & 
		\begin{tabular}{l}
		\textbf{We consider 15 radius features defined by the shortest distance from the region to the	following  types of POI:}\\
		Hospital, Clinic, College, School, Bus Stop, Subway Station, Airport, Train Station, Coach Station, Shopping Mall,\\
		Supermarket, Market, Shop, Police Station, Scenic Spot.
		\end{tabular}\\
        \midrule
	    \begin{tabular}{c}
	    \rule{0pt}{10pt}
	    Index of\\
	    Basic Living Facility
	    \end{tabular} & 
		\begin{tabular}{l}
		\textbf{This binary feature will be assigned one if there are all the following types of living facilities within $1km$:}\\
		Medical Service, Shopping Place, Sports Venue, Education Service, Food Service, Financial Service, Communication\\ Service, Public Security Organ and Transportation Facility
		\end{tabular}\\
		\bottomrule
	\end{tabular}
\end{table*}
\subsection{Baseline Descriptions and implementations}
\label{apdx-baseline}
Here we introduce the description and implementation of all the comparing baselines in details.
\begin{itemize}[leftmargin=*]
    \item \textbf{MLP}. The multi-layer perceptron (MLP) is a most classic artificial neural network (ANN), which stacks several fully connected layers to transform the input features. Specifically, we apply two fully connected layers for POI and image representation learning, respectively. Then the two representation vectors are fused by concatenation and fed into a LR classifier for urban village detection.
    \item \textbf{GCN}\cite{kipf2016semi}. Graph convolution network (GCN) is a classic message passing graph neural network, which aggregates features from neighboring nodes based on the adjacency matrix. In our urban village detection problem, we first apply the dimension reduction for image features. And then, considering the gap between different modalities, we adopt two 2-layer graph convolution layers for modality-wise representation learning, and fuse the multi-modal representation with an additional linear transformation before feeding them into the predictor.
    \item \textbf{GAT}\cite{velivckovic2018graph}. Graph attention network (GAT) is also a popular graph neural network using attention mechanism to learn proper weights for neighboring nodes in features aggregation. The implementation of GAT model is similar to that of GCN, with the only change of aggregation function.
    \item \textbf{ImGAGN}\cite{qu2021imgagn}. This model is a state-of-the-art method for imbalanced network embedding, which is consistent with the class distribution of urban village detection (UV regions are in the minority class while non-UV regions are in the majority class). ImGAGN adopts adversarial learning to generate a set of synthetic minority nodes and balance the different classes. In our experiments, we adopt 3-layer MLP as synthetic minority nodes generator with the hidden size recommended in \cite{qu2021imgagn}. As for the two important predefined parameters of this model (i.e. the training minority nodes ratio $\lambda_1$ and discriminator training steps $\lambda_2$), we respectively set them as $\lambda_1 = 1.0$ and $\lambda_2 = 100$, which has been proved to achieve the best performance in \cite{qu2021imgagn}.
    \item \textbf{MMRE}\cite{jenkins2019unsupervised}. Multi-modal Region Encoder (MMRE) is a state-of-the-art graph convolution based approach to address the Learning an Embedding Space for Regions (LESR) problem with multi-modal data. It defines a discriminator function to unify the POI features and satellite image features for region embedding. For the implementation of MMRE, we adopt three fully connected layers with hidden size 120, 84, 64 as encoder and a symmetry structure as decoder to constitute the denoising autoencoder for image representation learning. Meanwhile, a 2-layer GCN with 128 and 64 hidden units are used to learn POI representation. The SkipGram loss used for learning how to distinguish true contextual regions are calculated by 4 positive samples and 10 negative samples. The trade-off hyper-parameters in final training objective are set to $\lambda_I=0.5$ and $\lambda_s=0.1$ for autoencoder reconstruction loss and SkipGram loss, respectively. We remove the transition reconstruction loss since we do not use taxi mobility flow data in this work.
    \xcxre{\item \textbf{MUVFCN}\cite{mast2020mapping}. This is a state-of-the-art method for urban village detection. It adopts the fully convolutional neural network (FCN) with the pre-trained VGG19 \cite{simonyan2014very} model as the backbone. In our experiments, we implement it with FCN-8s architecture \cite{long2015fully}, and the average pooling is applied on output maps to obtain a 32-dimensional feature vector for final prediction.}
    \item \textbf{UVLens}\cite{chen2021uvlens}. This is a state-of-the-art method for urban village detection. It uses taxi trajectories to segment the city-wide satellite image into patches. Then, it integrates bike-sharing drop-off data into image patches and adopts Mask-RCNN \cite{he2017mask} model to detect UVs. In our experiments, due to the unavailability of bike-sharing drop-off data, we use satellite images for UV detection. We first exploit a histogram equalization as recommended in \cite{chen2021uvlens} and adopt CNN backbone to extract feature maps for regions. Since we have divided the urban area into grids of fixed size, these grids can be treated as positive candidate object bounding boxes. Thus, we omit Region Proposal Network (RPN) as well as ROIPooling \cite{he2017mask}, and directly extract high-level semantic features from feature maps with stacked fully connected layers of 4096, 4096, 128 and 64 hidden units for final prediction.
\end{itemize}

\subsection{POI Feature Construction}
\label{adpx-poi}
As mentioned in section \ref{poi-features}, we construct three groups of POI features to describe the basic living facilities of regions, which are category distribution, POI radius and the index of basic living facility. Here we list all the POI types considered in our work in Table \ref{poi_type}. Note that the types of POIs used to define the index of basic living facility are mainly selected according to an official document released by Ministry of Housing and Urban-Rural Development of China \footnote{http://www.mohurd.gov.cn/wjfb/201811/W02018113004480.pdf}. 

\subsection{Ground-truth Collection}\label{apx:gtc}
In our work, we exploit crowdsourcing to collect ground-truth UV and non-UV regions for the three real-world datasets with the following steps.
We first make great efforts to collect the news reports and official documents related to urban village (such as urban village renovation and demolition plans) on the Internet, based on which we obtain a set of potential UVs to be verified. 
Then, we recruit a group of professional participants to pick out the region grids that they think are certainly contained by or significantly overlapped with UVs on an online crowdsourcing platform. The platform provides the geographical coordinates of each candidate region with embedded online maps. The participants investigate these regions through satellite images and street views with map service for determining whether they are UV regions or not.
To obtain more reliable labeled data, we assign each region to 3 participants, and the region will be labeled as UV only if all three participants reach consistency. As for non-UVs, we randomly sample a number of residential areas and ask the participants to check these regions in the same way.

\end{document}